\documentclass[nohyperref]{article}

\usepackage{microtype}
\usepackage{graphicx}
\usepackage{subfigure,nicefrac}
\usepackage{booktabs} 

\usepackage{hyperref}

\usepackage[accepted]{icml2022}

\usepackage{amsmath}
\usepackage{amssymb}
\usepackage{mathtools}
\usepackage{amsthm}

\usepackage[capitalize,noabbrev]{cleveref}

\theoremstyle{plain}
\newtheorem{theorem}{Theorem}[section]

\newtheorem{lemma}[theorem]{Lemma}

\theoremstyle{definition}

\newtheorem{assumption}[theorem]{Assumption}
\theoremstyle{remark}

\theoremstyle{fact}
\newtheorem{fact}[theorem]{Fact}

\usepackage[textsize=tiny]{todonotes}

\icmltitlerunning{On Finite-Sample Identifiability of Nonlinear ICA}

\usepackage{microtype,tikz,bm}
\usepackage{graphicx}
\usepackage{amsmath}

\usepackage{amsfonts}
\usepackage{amssymb}
\usepackage{booktabs,epstopdf} 
\usepackage{amsthm}
\usepackage{wrapfig}
\usepackage{lipsum}
\usepackage{pgfplots}
\usepackage{pgfplotstable}
\usepackage{graphicx} 
\usepackage{caption}
\usepackage{comment}
\usepackage{times} 
\usepackage{newtxmath,bbold}
\usepackage{xcolor}

\newcommand{\W}{\boldsymbol{W}}

\newcommand{\Q}{\boldsymbol{Q}}

\newcommand{\A}{\boldsymbol{A}}

\newcommand{\s}{\boldsymbol{s}}

\newcommand{\x}{\boldsymbol{x}}

\newcommand{\z}{\boldsymbol{z}}

\newcommand{\y}{\boldsymbol{y}}
\renewcommand{\u}{\boldsymbol{u}}

\renewcommand{\a}{\boldsymbol{a}}

\newcommand{\T}{{\!\top\!}}

\usepackage{xcolor}
\usepackage{psfrag,framed}
\usepackage{lipsum}
\usepackage{chapterbib}
\PassOptionsToPackage{normalem}{ulem}
\usepackage{ulem}
\definecolor{shadecolor}{RGB}{220,220,220}
\usepackage{graphicx}

\definecolor{orange}{RGB}{255,107,0}

\usepackage{threeparttable}

\begin{document}

\twocolumn[
\icmltitle{On Finite-Sample Identifiability of Contrastive Learning-Based Nonlinear Independent Component Analysis}



\icmlsetsymbol{equal}{*}

\begin{icmlauthorlist}
\icmlauthor{Qi Lyu}{yyy}
\icmlauthor{Xiao Fu}{yyy}
\end{icmlauthorlist}

\icmlaffiliation{yyy}{School of EECS, Oregon State University, Corvallis, OR, United States}

\icmlcorrespondingauthor{Xiao Fu}{xiao.fu@oregonstate.edu}
\icmlcorrespondingauthor{Qi Lyu}{lyuqi@oregonstate.edu}

\icmlkeywords{Machine Learning, ICML}

\vskip 0.3in
]



\printAffiliationsAndNotice{}  

\begin{abstract}
{ 
Nonlinear independent component analysis (nICA) aims at recovering statistically independent latent components that are mixed by unknown nonlinear functions. Central to nICA is the identifiability of the latent components, which had been elusive until very recently.
Specifically, Hyv{\"a}rinen {\it et al.} have shown that the nonlinearly mixed latent components are identifiable (up to often inconsequential ambiguities) under a generalized contrastive learning (GCL) formulation, given that the latent components are independent conditioned on a certain auxiliary variable.
The GCL-based identifiability of nICA is elegant, and establishes interesting connections between nICA and popular unsupervised/self-supervised learning paradigms in representation learning, causal learning, and factor disentanglement. However, existing identifiability analyses of nICA all build upon an unlimited sample assumption and the use of ideal universal function learners---which creates a non-negligible gap between theory and practice.  
Closing the gap is a nontrivial challenge, as there is a lack of established ``textbook'' routine for finite sample analysis of such unsupervised problems.
This work puts forth a finite-sample identifiability analysis of GCL-based nICA. Our analytical framework judiciously combines the properties of the GCL loss function, statistical generalization analysis, and numerical differentiation. 
Our framework also takes the learning function's approximation error into consideration, and reveals an intuitive trade-off between the complexity and expressiveness of the employed function learner.
Numerical experiments are used to validate the theorems.
}

\end{abstract}

\section{Introduction}
Independent component analysis (ICA) has been an indispensable unsupervised learning tool across multiple domains. Theory and methods have been developed for ICA since the 1990s; see, e.g., \cite{Comon1994}. The classic ICA guarantees to identify linearly mixed statistically independent latent components in an unsupervised fashion. The ICA technique has advanced many tasks such as blind speech/audio separation and brain signal denoising. Since the late 1990s, attempts have been made towards generalizing the classic  linear mixture model (LMM)-based ICA to nonlinear mixture models, e.g., in \cite{Hyvarinen1999,taleb1999source,ziehe2003blind,oja1997nonlinear}, driven by the ubiquity of nonlinearity in real-world data.

Formally, the {\it nonlinear independent component analysis} (nICA) problem deals with scenarios where statistically independent latent components are mixed by {\it unknown} nonlinear functions. 
The nICA technique aims at recovering the latent components up to certain (inconsequential) ambiguities. The nICA task finds many connections to modern machine learning and unsupervised representation learning problems. For example, a number of works used the nICA perspective to develop deep neural feature extractors for latent factor disentanglement  \cite{bengio2013representation,locatello2019challenging,higgins2016beta,kim2018disentangling,chen2018isolating}. The disentanglement perspective was further connected to causal factor learning \cite{peters2017elements,zhang2009identifiability,monti2020causal}.
Furthermore, nICA and its close relatives were also used to understand popular neural representation learning frameworks such as contrastive learning \cite{hyvarinen2016unsupervised,hyvarinen2017nonlinear,hyvarinen2019nonlinear}, variational autoencoder (VAE) \cite{khemakhem2020variational} and data-augmented self-supervised learning \cite{zimmermann2021contrastive}. 
For all these tasks, nICA offers theory-driven perspectives to understand their successes and sometimes to improve their learning methods. In particular, the {\it identifiability} of the latent independent components under nICA models can often provide useful insights into these aspects.

The (n)ICA identifiability problem is {concerned with the conditions and learning criteria under which} one can reverse the unknown mixing process to recover the latent components.
However, unlike the classic LMM-based ICA whose model identifiability is well-studied \cite{Comon1994, hyvarinen2000independent,Common2010}, identifiability of nICA models had not been fully understood for a long period. In fact, it is well-known that an nICA model that only assumes statistical independence of the latent components is not identifiable \cite{hyvarinen1999nonlinear}.

In recent years, a number of new nICA paradigms emerged, which nicely addressed the identifiability challenge under some additional yet physically meaningful conditions. These paradigms judiciously utilized structural information about the latent components, e.g., temporal dependence \cite{sprekeler2014extension,hyvarinen2017nonlinear} and non-stationarity \cite{hyvarinen2016unsupervised} of data, to underpin the model identifiability. In particular, the work in \cite{hyvarinen2019nonlinear} unified these developments under a {\it generalized contrastive learning} (GCL) based framework \cite{gutmann2010noise}, where the latent components are assumed to be conditionally independent given an auxiliary variable. 
Under the GCL framework, the latent components are identifiable up to component-wise invertible nonlinear transformations, by simply learning a logistic regression neural discriminator.

The surprising connection between contrastive learning and nICA is both refreshing and insight-revealing. The identifiability proofs in \cite{hyvarinen2016unsupervised,hyvarinen2017nonlinear,hyvarinen2019nonlinear} are also elegant. 
Nevertheless, a caveat is that the GCL framework, same as other identifiable nICA works \cite{khemakhem2020variational,locatello2020weakly,gresele2020incomplete}, assumes that unlimited data samples are available. This presents a non-negligible gap between nICA identifiability theory and practice, since one never has unlimited data in real systems.
{In addition, the GCL-based nICA works all assume that universal exact function learners are used in their learning process. However, function approximation errors always exist in practice, even if very expressive function learners, e.g., deep/wide neural networks, are used.}
These less realistic assumptions naturally lead to an inquiry: Can the identifiability of GCL-based nICA be established under limited sample cases {in the presence of learning function approximation errors}?

Filling this theory-practice gap is a nontrivial task. First, the proofs in \cite{hyvarinen2019nonlinear,hyvarinen2016unsupervised,hyvarinen2017nonlinear} heavily rely on the equivalence between the optimal logistic regressor and the log-probability density function (log-PDF) difference of the two contrastive classes {\it in the limit of infinite samples}.
Second, many steps in the proofs use first-order and second-order derivatives with respect to the learned latent components---whose existence over continuous open domains is also a result of the (uncountably) unlimited data assumption. In addition, unlike supervised learning where well-established generalization analysis routines (see, e.g., \cite{shalev2014understanding}) can be used to characterize finite sample performance, there is no such toolkit for latent component analysis. This is perhaps because supervised learning's success is measured by the ``distance'' between an empirical loss and its population version, but latent component identification often has a much more intricate objective---which varies across different generative models and learning goals.

\subsection{Contributions}
In this work, our interest lies in offering a finite-sample analysis for the GCL-based nICA framework.
Our analytical framework consists of three major steps.
First, we use the notion of restricted strong convexity of the logistic loss used in GCL to characterize the relationship between its optimal solution and gradient at the optimum. Second, we combine statistical generalization theorems with this relation to characterize the gap between the regressor learned from finite samples and {the optimum under the population case}. Third, based on the gap, we characterize the separability of different latent components using numerical differentiation tools. As a result, we show that GCL-based nICA can separate different latent components to a reasonable extent under finite samples, and the performance improves when the sample size grows. 
Our result also takes the learning function's approximation error into consideration, and reveals an intuitive trade-off between the complexity and expressiveness of the employed learning function.
To our best knowledge, the result is the first to establish such a finite sample identifiability under the GCL-based nICA framework.
We also envision that our proof technique could help understand the finite-sample performance of different but nICA-related frameworks, e.g., those in \cite{zimmermann2021contrastive, khemakhem2020variational,locatello2020weakly}.

\subsection{Notation}
We use the following notations: $x,\bm x,\bm X$ represent a scalar, vector, and matrix, respectively; $f'$ and $f''$ denote the first-order and second-order derivatives of function $f$, respectively; $f \circ g$ denotes the function composition of $f$ and $g$; $\sigma_{\min}(\bm{W})$ denotes the smallest nonzero singular value of matrix $\bm{W}$; $\mathbb{E}[\cdot]$ denotes expectation of its argument; $p(x)$ denotes the probability density function of random variable $x$; a column vector $\bm{a}\in\mathbb{R}^D$ is defined as $\a=[a_1,\ldots,a_D]^\T=(a_1,\ldots,a_D)$. 

We will frequently use the notation $\x_\ell\in {\cal X}$ to represent the $\ell$th sample drawn from a distribution ${\cal D}$ defined over the continuous domain ${\cal X}$. The notation $\x$ without subscript represents a random vector defined over the continuous domain ${\cal X}$ following the same distribution ${\cal D}$---i.e., $\x_\ell$ can be considered as the $\ell$th realization of the random vector $\bm{x}$.

\color{black}

\section{Background}

\subsection{ICA, nICA, and Model Identifiability}
The classic ICA techniques deal with the LMM, i.e.,
\begin{align}\label{eq:linear_ica}
    \bm{x}_\ell = \bm{A}\bm{s}_\ell,~\ell=1,\ldots,N,
\end{align}
where
$\bm{x}_\ell\in\mathbb{R}^M$ is the $\ell$th observed sample, $\bm{s}_\ell\in\mathbb{R}^D$ are the $D$ latent components, and $\bm{A}\in\mathbb{R}^{M\times D}$ with $M\geq D$.
It is assumed that $\x_\ell$ and $\bm s_\ell$ are the $\ell$th realizations of the random vectors $$\x=[x_1,\ldots,x_M]^\T,~\text{and}~\bm s=[s_1,\ldots,s_D]^\T,$$ respectively, in which $\x=\A\s$ and $s_1,\ldots,s_D$ are statistically independent. The task of ICA is to recover $\bm s_\ell$ from $\x_\ell$. Recovering the latent components from LMMs is in general not possible, since $\x_\ell=\A\s_\ell=\A\Q\Q^{-1}\s_\ell$ for any nonsingular $\Q$. However, using the statistical mutual independence among $s_1,\ldots,s_D$, one can show that $\s_\ell$ and $\A$ are identifiable through ICA techniques up to permutation and scaling ambiguities. This is normally done by finding an inverse filter $\W$ such that the elements of $\y=\W\x$ are mutually independent; see \cite{Comon1994, hyvarinen2000independent,sanjeev2012provable}.

In nICA, the LMM in \eqref{eq:linear_ica} is generalized to a nonlinear mixture model, i.e., \cite{hyvarinen1999nonlinear}
\begin{align}\label{eq:general_ica}
    \bm{x}_\ell = \bm{g}(\bm{s}_\ell),
\end{align}
where $\bm g(\cdot)$ is a smooth and invertible {\it unknown function}, and $\bm x_\ell$ and $\bm s_\ell$ are defined as before. The goal often amounts to learning a nonlinear function $\bm h(\cdot)$ such that for any $\x=\bm{g}(\s)$ the following holds:
\begin{align}\label{eq:ident_nica}
    \y&=\bm h(\x)~{\rm s.t.}~
    y_i=\sigma_i(s_{\pi(i)}),\quad i=1,\ldots,D,
\end{align}
where $\{\pi(1),\ldots,\pi(D)\}$ represents a permutation of $\{1,\ldots,D\}$ and $\sigma_i(\cdot):\mathbb{R}\rightarrow \mathbb{R}$ is an unknown invertible function. Note that such $y_i$ and $s_{\pi(i)}$ attain the maximum mutual information and can be converted from one to another. Thus, the learning goal is meaningful.
Unfortunately, unlike ICA, under such a nonlinear mixture model, the desired $y_i$ in \eqref{eq:ident_nica} is in general not identifiable by just constraining the output of a learning system to be statistically independent \cite{hyvarinen1999nonlinear}

\subsection{Auxiliary Variable-Assisted GCL-based nICA}\label{sec:aux}

In recent years, some notable breakthroughs of the identifiability research of nICA have been made.
Specifically, several recent works show that $\bm s$ in \eqref{eq:general_ica} can be identified (i.e., \eqref{eq:ident_nica} can be guaranteed) under interesting and physically meaningful conditions; see, e.g., \cite{hyvarinen2019nonlinear,hyvarinen2017nonlinear,hyvarinen2016unsupervised}. In particular, \cite{hyvarinen2019nonlinear} distilled the essence and presented a unified framework based on GCL. Under the framework, it is assumed that $s_1,\ldots,s_D$ are statistically independent {\it conditioned on} the revelation of an {\it auxiliary variable} $\bm{u}$, i.e.,
\begin{align}\label{eq:auxiliary}
    \log p(\bm{s}|\bm{u}) = \sum_{i=1}^D q_i(s_i,\bm{u}),
\end{align}
where $q_i(\cdot)$ is a certain continuous function and $p(\bm{s}|\bm{u})$ is the conditional PDF of $\s$ given $\bm u$. Note that $\u$ is observed together with $\x$, i.e., $\z=(\x,\u)$ appears as a pair.

To put into context, we briefly mention some examples where the existence of auxiliary variables makes sense:

{
\paragraph{Example - Time Contrastive Learning (TCL)} In TCL, the auxiliary variable could be an indicator of the time stamp with $u=t$ or other information, e.g., the mean and variance of of $\s_\ell$ in a certain time frame \cite{hyvarinen2016unsupervised,hyvarinen2017nonlinear,hyvarinen2019nonlinear}. Here, the condition \eqref{eq:auxiliary} means that the latent variables are independent if they are from the same time frame/slot.

\paragraph{Example - Multiview Contrastive Learning (MVCL)} A slightly different but closely related example is MVL. There, $\bm{s}$ is assumed to be mixed with different nonlinear functions for each view \cite{gresele2020incomplete}, i.e, $\bm{x}=\bm{g}(\bm{s})$ with the auxiliary variable $\bm{u}\approx \widetilde{\bm{g}}(\bm{s})$ that is another view of the same data entity with some random perturbations. In this case, the log PDF $\log p(\bm{s}|\bm{u})$ can also be similarly factored as in \eqref{eq:auxiliary} given that $s_1,\ldots,s_D$ are statistically independent. 

}

Under \eqref{eq:auxiliary},
the nICA framework in \cite{hyvarinen2019nonlinear}
proposed to learn a regression function
\begin{align}\label{eq:regression}
    {r}(\bm{x},\bm{u}) = \sum_{i=1}^D \phi_i\left(h_i\left(\bm{x}\right),\bm{u}\right)
\end{align}
to distinguish between two types of $\z_\ell$, i.e.,
\begin{equation}\label{eq:pnsamples}
     \z_\ell=(\bm{x}_\ell,\bm{u}_\ell) \text{ and } \z_\ell=(\bm{x}_\ell,\widetilde{\bm{u}}_\ell).
\end{equation}
Note that the ``positive samples'' $\z_\ell=(\bm{x}_\ell,\bm{u}_\ell)$ are observed from data---and $\u_\ell$ is the natural auxiliary variable of $\x_\ell$. 
However, for the ``negative samples'' $\z_\ell=(\bm{x}_\ell,\widetilde{\bm{u}}_\ell)$, $\widetilde{\bm{u}}_\ell$ is randomly drawn from $p(\bm{u})$ which has no dependence on $\bm{x}_\ell$. 
For example, in TCL, the positive sample $\z_\ell=(\bm{x}_\ell,\bm{u}_\ell)=(\bm{x}_\ell,t)$ where $\ell\in {\rm time~frame~}t$, but the negative sample $\z_\ell=(\bm{x}_\ell,\widetilde{\bm{u}}_\ell)=(\bm{x}_\ell,t')$ where $\ell\notin{\rm time~frame~}t' $.
In MVCL, the positive sample $\z_\ell=(\bm{x}_\ell,\bm{u}_\ell)$ which means that the two views are generated from the same $\s_\ell$.
The negative sample $\z_\ell=(\bm{x}_\ell,\widetilde{\bm{u}}_\ell = \u_j)$ with $\ell\neq j$; i.e., the pair of data from the two views correspond to different latent vectors $\s_\ell$ and $\s_j$.

\begin{figure}[t!]
    \centering
	\subfigure{\includegraphics[trim=13.0cm 6.5cm 2.2cm 0, clip,width=1\linewidth]{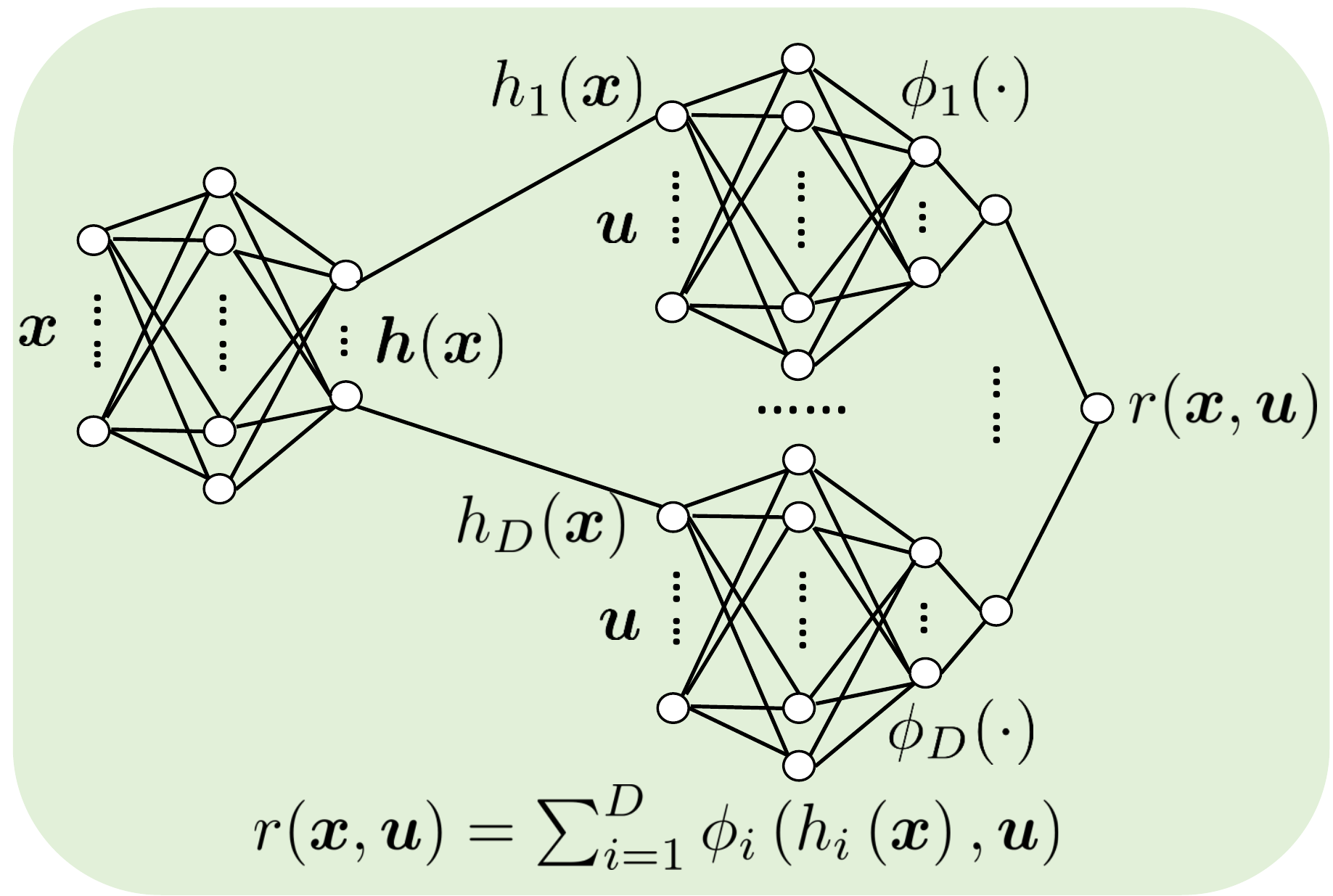}}
	\caption{The neural network structure used for GCL-based nICA \cite{hyvarinen2019nonlinear}.}\label{figs:net_struc}
\end{figure}

The functions $h_i$ and $\phi_i$ are often represented by nonlinear function learners, e.g., neural networks. This ``classification problem'' can be realized using a logistic loss:
\begin{align}\label{eq:gcl}
    \min_{\bm{\phi},\bm{h}}~\mathcal{L}=\min_{\bm{\phi},\bm{h}}~\mathbb{E}_{\bm z} \left[\log(1+\exp[-d {r}(\bm z)])\right],
\end{align}
where $d\in \{+1,-1\}$ is the ``label'' of $\z$. The realizations of $d$, namely, $d_\ell$ for $\ell=1,\ldots,N$, are created using the following rule:
\begin{equation}
    d_\ell =\begin{cases}  +1,&\quad \z_\ell=(\x_\ell,\u_\ell),\\
    -1,&\quad \z_\ell=(\x_\ell,\widetilde{\u}_\ell).\end{cases}
\end{equation}
The learning system is shown in Fig.~\ref{figs:net_struc}.

The framework is called ``constrastive learning'' because it constructs a discriminator from data itself without using traditional class labels as in supervised learning. 
Similar ideas are widely used in the domain of self-supervised representation learning \cite{chen2020simple,he2020momentum,tian2020contrastive,oord2018representation}.

The work in \cite{hyvarinen2019nonlinear} showed that the learned $\bm h^\star(\x)$ (i.e., the optimal solution of \eqref{eq:logistic_population} in Appendix~\ref{ap:proof_r_gap}) is the desired latent component $\s$ up to some ambiguities. To see the result,
let us define the following
\begin{align*}
\bm{y} =\bm{h}(\bm{x}),\quad\bm{v}(\bm{y}) =\bm{g}^{-1}(\bm{h}^{-1}(\bm{y}))=\bm{s}.
\end{align*}
Using the above notations, we restate the main result of \cite{hyvarinen2019nonlinear} as follows:

\begin{theorem}[{\bf Infinite-Sample Identifiability}]\label{thm:general_infinite}\cite{hyvarinen2019nonlinear}
    Assume 
    
     (i) that the data follows the model in \eqref{eq:general_ica} and \eqref{eq:auxiliary} with $M=D$; the conditional log-PDF $q_i$ in \eqref{eq:auxiliary} is smooth (i.e., second-order differentiable) as a function of $s_i$ for any $\bm{u}$;
        
     (ii)    ({\bf Variability Assumption}) that for any $\bm{y}\in\mathbb{R}^D$, there exist $2D+1$ vector $\bm{u}_j$'s, such that the $2D$ vectors in $\mathbb{R}^{2D}$ denoted as
        \begin{align}\label{eq:def_W}
            \bm{W}&=[\bm{w}(\bm{y},\bm{u}_1)-\bm{w}(\bm{y},\bm{u}_0),\cdots, \nonumber\\
            &\quad\quad\bm{w}(\bm{y},\bm{u}_{2D})-\bm{w}(\bm{y},\bm{u}_0)]
        \end{align}
        are linearly independent, where
        \begin{align}\label{eq:def_w_vector}
            \bm{w}(\bm{y},\bm{u})=&\left[\frac{\partial q_1(y_1,\bm{u})}{\partial y_1},\cdots,\frac{\partial q_D(y_D,\bm{u})}{\partial y_D},\right.\\
            &\ \ \left.\frac{\partial^2 q_1(y_1,\bm{u})}{\partial y_1^2},\cdots,\frac{\partial^2 q_D(y_D,\bm{u})}{\partial y_D^2}\right]; \nonumber
        \end{align}
        
       (iii)  that \eqref{eq:gcl} is solved with universal function approximators to represent $r(\z)$; 
       
       (iv) and that the learned optimal $ \bm h^\star = (h_1^\star,\cdots,h_D^\star)$ is constrained to be invertible and smooth.
       
    Then, in the limit of infinite data, we have
    $
        h^\star_{\pi(i)}(\bm{x}) = v^{-1}_i(s_i),
    $ for $i=1,\ldots,D$, where $\{\pi(1),\ldots,\pi(D)
    \}$ is a permutation of $\{1,\ldots,D\}$.
\end{theorem}
The variability assumption means that the auxiliary variable $\bm{u}$ must provide sufficiently different information and impacts on the independent components to identify---i.e., the realizations of $\u$ should be diverse; see more explanations in \cite{hyvarinen2019nonlinear,gresele2020incomplete}.

The proof of Theorem~\ref{thm:general_infinite} consists of three major steps. {{\bf Step 1}: Under the unlimited data assumption, the optimal logistic regression function is converged to the log-density difference of the two classes \cite{goodfellow2014generative}.
To be specific, one can show that
\begin{align}\label{eq:logpdfdiff}
   {r}^\star(\bm{x},\bm{u})&= \sum_{i=1}^D\phi_i^\star \left(h_i^\star\left(\bm{x}\right),\bm{u}\right) \nonumber\\
   & = \log p(\bm{x} | \bm{u})  - \log p(\bm{x}),
\end{align}
if $r^\star$ is learned using a universal function approximator.

{\bf Step 2}: Using the assumption in \eqref{eq:auxiliary}, a functional equation can be established everywhere over ${\cal X}$ (i.e., the domain of $\x$) by equating the constructed \eqref{eq:regression} and \eqref{eq:logpdfdiff}.

{\bf Step 3}: Given the equation holds everywhere, cross-derivatives w.r.t. $y_j$ and $y_k$ are taken, which results in a linear system with a full-rank coefficient matrix under the assumption of Variability---which finally leads to the desired result. } 

We have restated the proof in Appendix~\ref{ap:proof_infinite}, as it may help the readers better understand the finite-sample analysis.

One can see that all the three steps heavily rely on the unlimited sample assumption. 
In the next section, we will offer an analytical framework that can circumvent this unrealistic assumption.

\section{Finite-Sample Analysis of nICA}
In practice, instead of directly solving \eqref{eq:gcl}, one always deals with the corresponding empirical loss function as follows:
\begin{align}\label{eq:gcl_emp}
    \min_{\bm{\phi},\bm{h}}~\widehat{\mathcal{L}}=\min_{\bm{\phi},\bm{h}}~\frac{1}{N}\sum_{\ell=1}^N \log\left(1+\exp\left[-d_\ell {r}(\z_\ell)\right]\right).
\end{align}

Before stating the main results, we first define the vector that we hope to characterize as
\begin{align}\label{eq:gamma}
    \bm{\gamma}_{jk}=&\left[
    \frac{\partial^2 {v}_1(\bm{y})}{\partial y_j \partial y_k},\cdots, \frac{\partial^2 {v}_D(\bm{y})}{\partial y_j \partial y_k}
    \right]^\top.
\end{align}
The vector $\bm{\gamma}_{jk}$ will be used as a key metric to quantify the latent component identification performance.
{Fact~\ref{lemm:gamma}} shows the rationale behind using such a vector to serve as our success metric---i.e., in the population case, $\bm{\gamma}_{jk}=\bm{0}$ holds for all $(j,k)$ pairs where $j<k$ everywhere. 

{
\begin{fact}\label{lemm:gamma} 
    In the proof of Theorem~\ref{thm:general_infinite}, it is noted that if $\bm \gamma_{jk} =\bm 0$ for all $(j,k)$'s, then we have
    $
        h^\star_{\pi(i)}(\bm{x}) = v^{-1}_i(s_i),
    $ for $i=1,\ldots,D$, where $\{\pi(1),\ldots,\pi(D)
    \}$ is a permutation of $\{1,\ldots,D\}$.
\end{fact}}

{\bf Proof:} 
First note that $\bm{\gamma}_{jk}=\bm{0}$ means that
\begin{align}\label{ap_eq:cross_zero}
    \frac{\partial^2 {v}_i(\bm{y})}{\partial y_j \partial y_k} = 0
\end{align}
for any $i\in\{1,\cdots,D\}$. Since we have $\bm{v}(\bm{y}) =\bm{g}^{-1}(\bm{h}^{-1}(\bm{y}))$
where both $\bm{g}$ and $\bm{h}$ are smooth and invertible functions, $\bm{v}=\bm{g}^{-1}\circ\bm{h}^{-1}$ is also invertible, which leads to the fact that the Jacobian $\bm{J}_{\bm{v}}$ should be full-rank:
\begin{equation}\label{eq:Jv}
    {\rm rank}(\bm J_{\bm v})=D.
\end{equation}
Meanwhile, \eqref{ap_eq:cross_zero} implies that any $v_i$ only depends on one of its arguments $y_j$. That is, the Jacobian must have the following form
\[\bm{J}_{\bm{v}} = \text{Diag}(\bm{\lambda}) \bm{\Pi}\] where $\lambda_i\neq0$ for $i=1,\cdots,D$ and $\bm{\Pi}$ is a permutation matrix.

{Note that it is impossible that different $v_i$ and $v_{i'}$ are functions of the same $y_j$---otherwise the Jacobian would have at least a zero column, which violates ${\rm rank}(\bm J_v)=D$ in \eqref{eq:Jv}.}

As a result, we have $h^\star_{\pi(i)}(\bm{x}) = v^{-1}_i(s_i)$ for $i=1,\ldots,D$. \hfill $\blacksquare$

{Fact~\ref{lemm:gamma}} indicates that the ``size'' of $\|\bm \gamma_{jk}\|$ could quantify the level of success for separating the functions of $s_1,\ldots,s_D$.
Hence, under the finite sample scenario, our goal is to show that $\|\bm{\gamma}_{jk}\|$ is bounded by $O((\nicefrac{1}{N})^\beta)$ for a certain $\beta>0$.

To start with, we first characterize the Rademacher complexity of the neural network function class used to model the regression function $r(\z)$. We define the function class $\mathcal{F}$ for multi-layer perceptrons (MLP). 
{
\begin{assumption}[Neural Network]\label{as:nn}
    Assume that $\bm{h}(\cdot):\mathbb{R}^D\rightarrow\mathbb{R}^D$ and each $\phi_i(\cdot):\mathbb{R}^{D} \rightarrow \mathbb{R}$ is parameterized by an $L$-layer neural network with the following structure and constraint
    \begin{equation}\label{eq:hclass}
    \begin{aligned}
       \mathcal{F}=\{\bm{f}|\bm{f}(\bm{z})=\bm P_L\bm \zeta(... \bm P_2\bm \zeta(\bm P_1 \bm{z})),\ \|\bm P_i\|_F\leq B_i\}, 
   \end{aligned}
   \end{equation}
   where the activation function $\bm \zeta(\cdot)=[\zeta_1(\cdot),\ldots,\zeta_{D_j}(\cdot)]^\T$, and $\zeta_i(\cdot):\mathbb{R}\rightarrow\mathbb{R}$ is a 1-Lipschitz continuous function that satisfies $\zeta_i(0)=0$ for $i=1,\ldots,{D_j}$ for $j=1,\ldots,L$, and $D_j$ is the network width of the $j$th layer.
\end{assumption}
}

{We hope to remark that the identifiability theory in \cite{hyvarinen2019nonlinear} and this work require that $\bm h$ to be invertible. Such $\bm h$ could be approximated using special networks such as normalizing flows or autoencoder-type regularization. Nonetheless, we use a generic neural network following \cite{hyvarinen2019nonlinear}, which suggested that such simple constructions of $\bm h$ normally do not hurt the performance.}

We have the following bound for the complexity of $\mathcal{F}$.
\begin{lemma}[\cite{golowich2018size}, Corollary 1]
    Assume that the observation is bounded as $\|\bm{z}\|_2\leq C$. The Rademacher complexity $\mathfrak{R}^f_N$ of the function class defined in \ref{as:nn} is bounded by
    \begin{align}\label{eq:rademacher}
        &\mathfrak{R}^f_N \leq \\
        & {O}\left(C\prod_{i=1}^L B_i \min\left\{ \frac{\log^{3/4}(N)\sqrt{\log\left(\frac{C}{\Gamma}\prod_{i=1}^L B_i\right)}}{N^{1/4}}, \sqrt{\frac{L}{N}} \right\}\right),\nonumber 
    \end{align}
    where $\sup_{\bm{z}\in\mathcal{Z}}|r(\bm{z})|\geq\Gamma$.
\end{lemma}
The bound can be simplified to 
$\mathfrak{R}^f_N \leq O(C\prod_{i=1}^L B_i\sqrt{\nicefrac{L}{N}}).$

Under this simplification, we have the following complexity bound for $r(\z)$:
\begin{lemma}\label{lemm:rademacher}
    Assume that the observation is bounded as $\|\bm{x}\|_2\leq C_x$ and $\|\bm{u}\|_2\leq C_u$. Then the Rademacher complexity $\mathfrak{R}_N$ of $r(\z)$ is bounded by
    \begin{align*}
        \mathfrak{R}_N \leq O\left(\left[C_x \prod_{i=1}^L B_i+\sqrt{D} C_u\right]\prod_{i=1}^L B_i \sqrt{\frac{DL}{N}}\right).
    \end{align*}
\end{lemma}

The detailed proof is in Appendix~\ref{ap:proof_rademacher}. 

Under the population case, Step 1 in the proof of Theorem~\ref{thm:general_infinite} assumed that $r^\star(\bm{z})$ equals to the log of the PDF differences of the positive and negative classes {[cf. Appendix~\ref{ap:proof_infinite}]}. Then, the next steps can take derivatives using this equation over the continuous domain where $\z$ is defined. However, with $N$ samples, the distance between the learned regression function and the log-PDF difference is not zero. To characterize this distance, we introduce the following lemma:
\begin{lemma}\label{lemm:rsc}
    Assume that $|r(\z)|\leq \alpha$ over all $\z\in{\cal Z}$.
    The logistic function $\ell(r)=\log(1+e^{-d{r}})$ is $\gamma_\alpha$-restricted strongly convex ($\gamma_\alpha$-RSC) in $r$, where $\gamma_\alpha=\frac{e^{\alpha}}{(1+e^{\alpha})^2}$.
\end{lemma}

{\bf Proof:} Starting with the following function for any $(\bm{x},\bm{u})$
\begin{align*}
    \ell(r) = \log(1+e^{-d{r}}),
\end{align*}
we take derivative w.r.t. $r$, which gives us
\begin{align*}
    \frac{d \ell(r)}{d r} = \frac{-d}{1+e^{d r}},\ \frac{d^2 \ell(r)}{d r^2} = \frac{e^{d r}}{(1+e^{d r})^2}
\end{align*}
since $d=-1$ or $d=1$.

The function $\frac{e^{d r}}{(1+e^{d r})^2}$ is maximized when $r=0$ and it is monotonic for $r<0$ and $r>0$. By assuming that $|r|\leq \alpha$, we have
\begin{align*}
    \frac{e^{d r}}{(1+e^{d r})^2} &\geq \min\left\{\frac{e^{-\alpha}}{(1+e^{-\alpha})^2}, \frac{e^{\alpha}}{(1+e^{\alpha})^2}\right\}\\ 
    &= \frac{e^{\alpha}}{(1+e^{\alpha})^2}= \frac{e^{-\alpha}}{(1+e^{-\alpha})^2}
\end{align*}

Therefore, for bounded $r$, the logistic function is $\gamma_\alpha$-strongly convex, with
\begin{align*}
    \gamma_\alpha=\frac{e^{\alpha}}{(1+e^{\alpha})^2},
\end{align*}
which completes the proof. \hfill $\blacksquare$

Note that the restricted strong convexity of the logistic loss was often mentioned in the one-bit matrix/tensor recovery literature; see, e.g., \cite{ni2016optimal}.

{
In \cite{hyvarinen2019nonlinear}, it was assumed that $\widehat{r}^\star(\z)$ is a learning function that is an universal function approximator. Hence, $\widehat{r}^\star(\z)$ can exactly express the desired function [i.e., a log-PDF difference as on the right hand side of \eqref{eq:logpdfdiff}]}.
In practice, we take $r$ from a function class $\mathcal{R}$ that may not be powerful enough to express any function, even if ${\cal R}$ is a deep neural network class. To take this function mismatch into consideration, we make the following assumption:
\begin{assumption}\label{as:mismatch}
    Assume that we use $r\in \mathcal{R}$ as the learning function. The best learned $r\in\mathcal{R}$ from the population case \eqref{eq:gcl} is characterized as
    \begin{align}\label{eq:mismatch}
        \min_{r\in\mathcal{R}}\mathcal{L}(r)
        - \mathcal{L}(r^\star) \leq \nu
    \end{align}
    where $r^\star$ is the desired regression function as in \eqref{eq:logpdfdiff}.
\end{assumption}
Note that the bound $\nu$ serves as an indicator of the expressiveness of the function class ${\cal R}$. For example, when ${\cal R}$ consists of neural networks, $\nu$ decreases as the neural network becomes deeper and wider.

Using Lemma~\ref{lemm:rsc}, we show the following key lemma:
\begin{lemma}\label{lem:r_gap}
Assume that
    the empirical loss in \eqref{eq:gcl_emp} is trained with i.i.d. samples $\{\z_\ell\}_{\ell=1}^N$, and that the criterion is optimally solved. 
    Also assume that the solution of \eqref{eq:gcl_emp} is taken from the function  class ${\cal R}$.
    Then, we have the following bound over the domain $\z\in
    {\cal Z}=\mathcal{X}\times\mathcal{U}$ {with probability at least $1-\delta$}:
    \begin{align}
        &\mathbb{E}_{\cal D}[|\widehat{r}^\star(\bm{z})-r^\star(\bm{z})|^2]\\
        &\leq \frac{(1+e^\alpha)^2}{e^\alpha} \left(2\mathfrak{R}_N{+\nu}+5c\sqrt{\frac{2\ln(8/\delta)}{N}}\right),\nonumber
    \end{align}
    where $\x\in{\cal X}$ and $\u\in{\cal U}$ in which ${
    \cal X}$ and ${\cal U}$ are two continuous open domains, ${\cal D}$ is the distribution where any $\z_\ell\in{\cal Z}$ is drawn from,
    $\widehat{r}^\star$ and $r^\star$ are the optimal solutions of \eqref{eq:gcl_emp} and the desired regression function in \eqref{eq:logpdfdiff}, respectively,  $\alpha = \left(\sqrt{D}C_x\prod_{i=1}^L B_i+DC_u\right)\prod_{i=1}^L B_i$ is an upper bound of $\left|{r}(\bm{z})\right|$.
\end{lemma}

The proof of Lemma~\ref{lem:r_gap} can be found in Appendix~\ref{ap:proof_r_gap}.

The bound in Lemma~\ref{lem:r_gap} shows the expected distance between $\widehat{r}^\star$ and $r^\star$---and how the gap scales with various problem parameters.
This will be used in the next steps to estimate the numerical derivative in the proof of the main theorem:

\begin{theorem}[{\bf Main Result}]\label{thm:general_finite}
    Under the generative model \eqref{eq:general_ica} and Assumption \ref{as:nn}, assume that the learning problem defined in Theorem \ref{thm:general_infinite} is solved with $N$ i.i.d. samples $\{\bm{z}_\ell\}_{\ell=1}^N$, {and that the learned $\bm h$ is invertible.}
    Suppose that the learned $\widehat{r}^\star \in{\cal R}$ is fourth-order differentiable w.r.t. $\y$ and the absolute value of the fourth-order partial derivative of $t(\z)=\widehat{r}^\star\left(\bm z\right)-r^\star\left(\bm z\right)$ w.r.t. any $y_i$ is upper bounded by ${C_t}$.
    Then, we have the following bound with probability of at least $1-\delta$,
    \begin{align}\label{eq:mainresult}
        &\mathbb{E}_{\cal D}\left[\left\|\widehat{\bm{\gamma}}_{jk}\right\|_2^2\right] \\
        &\leq 
        O\left(\frac{D{C_t}(1+e^\alpha)}{e^{\alpha/2}{ \sigma_*^2}} \left(\mathfrak{R}_N+\nu+\alpha\sqrt{\frac{\ln(1/\delta)}{N}}\right)^{1/2}\right), \nonumber
    \end{align}
    where ${\cal D}$ is the distribution of $\z$, $\widehat{\bm{\gamma}}_{jk}$ is an estimation of $\bm \gamma_{jk}$ at any observed $\x_\ell$ using $N$ samples for any $(j,k)$ pair with $j < k$ where the upper bound $\alpha = (\sqrt{D}C_x\prod_{i=1}^L B_i+DC_u)\prod_{i=1}^L B_i$, and {$\sigma_*=\max\limits_{\W} \sigma_{\min}({\bm W})$ [cf. the definition of $\W$ in \eqref{eq:def_W}].}

\end{theorem}

Theorem \ref{thm:general_finite} asserts that with a large enough $N$, the $\widehat{\bm{\gamma}}_{jk}$'s are almost zero vectors for all $(j,k)$'s. As mentioned in {Fact~\ref{lemm:gamma}}, this serves as a quantified indicator for the latent component identification performance. 

The result in Theorem~\ref{thm:general_finite} is intuitive---it presents a trade-off between the learning function's complexity and its expressiveness.
Specifically, given a certain $N$, increasing the network complexity (e.g., by increasing the network depth and width) makes the learning function class ${\cal R}$ more expressive (i.e., with a smaller $\nu$) but more complex (i.e., with a larger $\mathfrak{R}_N$). If $N$ is not large, $\mathfrak{R}_N$ could dominate the right hand side of \eqref{eq:mainresult}.  Hence, under a small or moderate sample size, it is useful to employ a reasonably expressive network to serve as $\widehat{r}^\star$, but not encouraged to use an {\it overly} deep/wide neural network. This is similar to the widely recognized ``data-hungry'' and overfitting phenomena observed in supervised {\it deep} learning problems.

\paragraph{Proof Sketch.} We sketch the proof of Theorem~\ref{thm:general_finite} here---the readers are referred to the appendices for the detailed proof. Our proof consists of the following steps. 

First, starting from \eqref{eq:gcl_emp}, we estimate the performance of the learned regression function; i.e., we bound the following distance
\[ \mathbb{E}_
{\cal D}[|\widehat{r}^\star(\bm{z})-r^\star(\bm{z})|^2],\]
which can be done by invoking Lemma~\ref{lemm:rsc} and Lemma~\ref{lem:r_gap}.

Second, we construct
\begin{align}\label{eq:construct}
    {t(\y_\ell)} &= \sum_{i=1}^D q_i(v_i(\bm{y}_\ell),\bm{u}_\ell) - \log p_s(\bm{v}(\bm{y}_\ell)) \nonumber\\
    & -\sum_{i=1}^D \phi_i\left([\bm{y}_\ell]_i,\bm{u}_\ell\right).
\end{align}
at any observed sample $\ell$---which is a sample version of the key functional equation for establishing the infinite sample nICA identifiability; see more details in Appendix~\ref{ap:proof_infinite}.

Taking numerical derivatives of \eqref{eq:construct} w.r.t. $y_j$ and $y_k$, we establish a ``noisy'' system of linear equation $$\W{\bm \kappa}_{jk} = \bm b \approx \bm 0,$$ where $\|\bm b\|$ can be characterized by the quantity of $\mathbb{E}[|\widehat{r}^\star(\bm{z})-r^\star(\bm{z})|^2]$ and ${\bm \kappa}_{jk}$ includes $\widehat{\bm \gamma}_{jk}$ as a sub-vector.

Third, combining with the smallest singular value of $\W$, using standard perturbation analysis of system of linear equations can help estimate the upper bound of $\widehat{\bm \gamma}_{jk}$.  \hfill $\blacksquare$

The full proof is relegated to Appendix~\ref{ap:proof_main}.

\section{Related Works}
{The work in \cite{sanjeev2012provable} considered finite sample analysis for the classic ICA under the LMM. The recent work in \cite{lyu2021simplex} considered finite sample analysis of post-nonlinear mixture model. However, the post-nonlinear mixture model is a special kind of simplified nonlinear model, whose finite-sample analysis is much less challenging than the case in this work.
In \cite{lyu2022understanding,lyu2022finite}, sample complexity of latent component recovery was studied under the deep canonical correlation analysis and self-supervised learning settings. The learning criteria there are different from GLS, and are arguably easier to handle due to their least squares nature.
In contrastive learning, \cite{arora2019theoretical,tosh2021contrastive,tsai2020self,haochen2021provable} analyzed finite sample performance, but in terms of downstream classification error. This is more aligned with traditional generalization error analysis, instead of latent component identification as in this work.

We should mention that our analysis is built upon the nICA framework in Theorem 1 of  \cite{hyvarinen2019nonlinear}. There are also other nICA works in the literature. For example, \cite{zimmermann2021contrastive} showed that data-augmented contrastive learning can recover the latent components up to affine transformations, under some more specific conditions, e.g., the latent components' {inner product} follows a certain distribution. An exponential family distribution assumption on $\s$ was used in Theorem 3 of \cite{hyvarinen2019nonlinear}, which also helped connect nICA and VAE;} see \cite{khemakhem2020variational}. Our proof does not use assumptions on the distribution of $\s$.

\section{Numerical Validation}
In this section, we validate our theoretical results using synthetic and real data experiments.

\subsection{Synthetic Data -TCL }
\paragraph{Data Generation} 
We follow the setup of the first experiment in \cite{hyvarinen2019nonlinear} for TCL and consider time-domain data $\{\x_\ell\}_{\ell=1}^N$. We generate latent components $\s=[s_1,s_2]^\T\in\mathbb{R}^2$ that are divided into 5 different time frames. The latent component samples $[\s_\ell]_i$ for $i=1,2$ are generated using a distribution specified by $\u_\ell\in\{ \bm \omega_1,\ldots,\bm \omega_5 \}$; i.e., $\bm u_\ell$ are randomly drawn from 5 different vectors, each corresponding to a time frame. Specifically, $[\s_\ell]_i$ for $i=1,2$ and $\ell\in$ time frame $\tau$ are generated by the multiplication of a Gaussian distribution and a Laplacian distribution, and the mean and variance/scale information of the distributions are contained in $\bm \omega_\tau$. The multiplication is needed to meet the variability condition in Theorem~\ref{thm:general_infinite}; see more details in \cite{hyvarinen2019nonlinear}. The generative function $\bm{g}(\cdot):\mathbb{R}^{D}\rightarrow\mathbb{R}^D$ is a one-hidden-layer neural network with leaky ReLU activations. The network weights are drawn from the standard Gaussian distribution.
The vectors $\z_\ell=(\x_\ell,\u_\ell)$ for $\ell=1,\ldots,N$ can be considered as i.i.d. samples that are re-arranged into different time frames.
We run the TCL framework using different sample sizes $N$ with equally divided time frames.

\paragraph{Evaluation Metrics} 
The goal of nICA is to output $y_{\pi(i)} = v^{-1}_i(s_i)$ for $i=1,\cdots,D$. Hence, to evaluate the performance, we use the {\it mutual information} (MI) between the estimated $\widehat{y}_{\pi(i)}$ and the corresponding $s_i$ as our evaluation metric, since the MI of $\widehat{y}_{\pi(i)}$ and the associated $s_i$ is maximized when $\widehat{y}_{\pi(i)}=v_i^{-1}(s_i)$ with an invertible $v_i(\cdot)$.
Specifically, we estimate the MI using kernel density estimation \cite{kozachenko1987sample}. We compute the MI between each of the recovered $\widehat{y}_i$ and the ground truth $s_j$'s. 
Then, we use the Hungarian algorithm \cite{kuhn1955hungarian} to fix the permutation ambiguity.

\paragraph{Neural Network Settings} 
We model $\bm{h}(\cdot)$ and $\phi_i(\cdot)$ using three-hidden-layer neural networks. We test various $R$'s (i.e., the number of hidden neurons) for each layer, where $R\in\{4,8,16,32,64,128,256,512\}$. The activation function is ReLU. 
{Note that a larger $R$ means a more complex (wider) neural network, which has a higher expressive power \cite{hornik1991approximation,hassoun1995fundamentals} (i.e., a smaller $\nu$) but leads to a larger Rademacher complexity $\mathfrak{R}_N$.}
For optimization, we use the \texttt{Adam} optimizer \cite{kingma2014adam} with an initial learning rate $5\times 10^{-4}$.

\begin{figure}[t!]
	\centering
	\subfigure{\includegraphics[width=\linewidth]{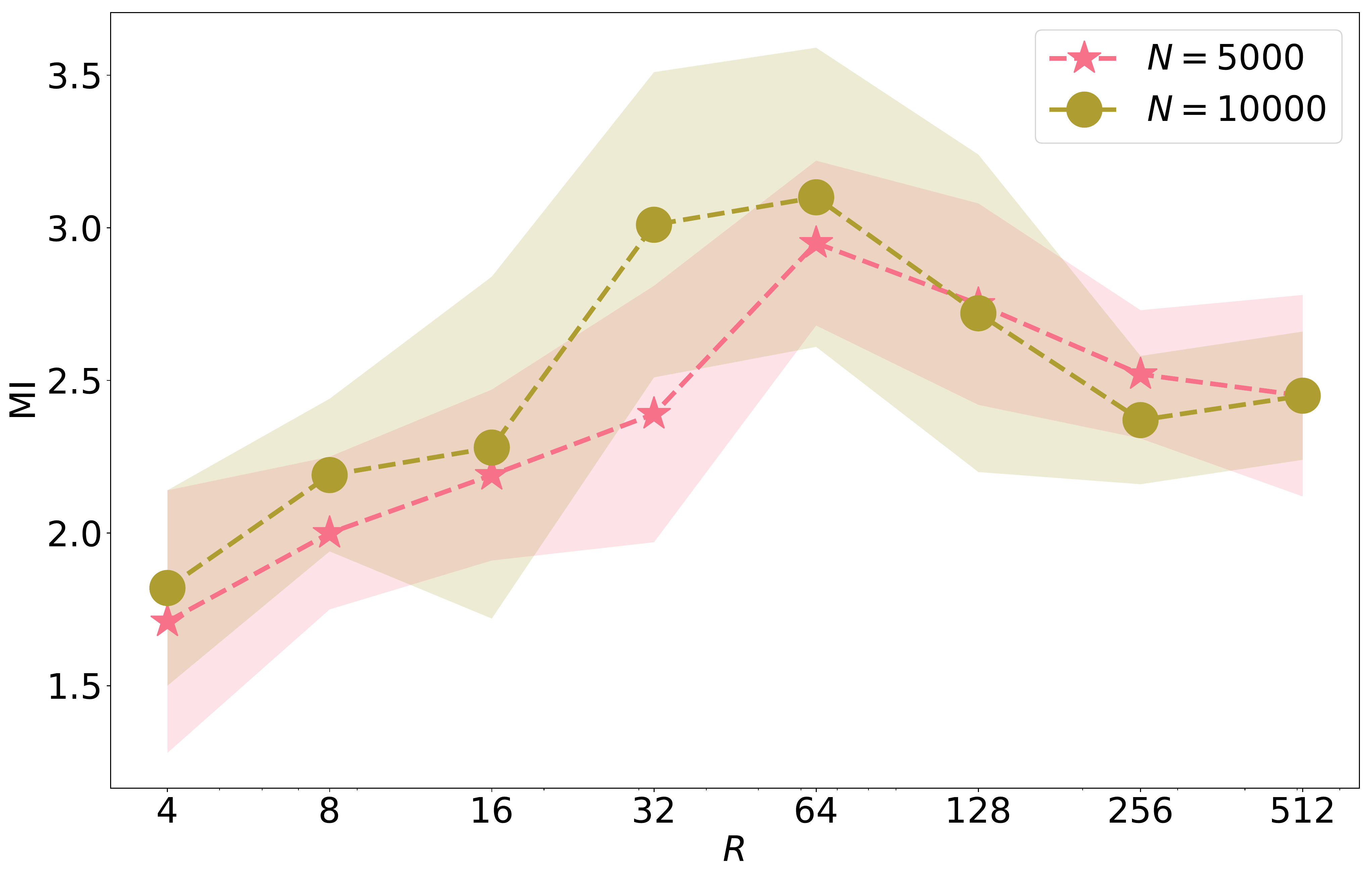}}
	\caption{The MI performance under the TCL setting.}\label{fig:synthetic_mi_contrastive}
\end{figure}

\paragraph{Results} Fig.~\ref{fig:synthetic_mi_contrastive} shows the nICA performance in terms of MI using different network width $R$'s under $N=5,000$ and $N=10,000$. The results are averaged over 5 random trials. 
One can see that, under a given $N$, the MI performance improves when the network size $R$ increases from 4 to 64. When $R$ continues to grow, the MI performance shows a descending trend. This exactly reflects the expressiveness ($\nu$) and complexity ($\mathfrak{R}_N$) trade-off revealed in Theorem~\ref{thm:general_finite}. That is, the initial performance improvement is likely due to the fact that wider neural networks can better approximate the desired unknown functions, and the decrease of MI may be due to the fact that the overly complex neural networks have a dominating $\mathfrak{R}_N$.

\subsection{Synthetic Data - MVCL}
\paragraph{Data Generation}
In this subsection, we use the multiview setting from \cite{gresele2020incomplete}---also see the second example in Sec.~\ref{sec:aux}. We use $\x_\ell=\bm g(\s_\ell)$ as the first view.
We set $\bm{u}_\ell=\widetilde{\bm g}(s_\ell)$ to be data from another view. We use $D=2$ with each component sampled from independent uniform distribution ${\rm U}[-a, a]$ with different $a$'s. For $\bm{x}$, the mixing function $\bm{g}$ is a one-hidden-layer neural network, with $D$ hidden neurons. For $\bm{u}_\ell$, the generation follows \cite{gresele2020incomplete} where $\bm{u}_\ell=\widetilde{\bm{g}}(\bm{s}_\ell+\bm{n}_\ell)$, in which $\bm{n}_\ell$ is again a product of a Gaussian variable and a Laplacian variable  \cite{hyvarinen2019nonlinear}.
Under this setting, the {\it sufficiently distinct views} condition in \cite{gresele2020incomplete} (which is derived from the variability assumption in Theorem~\ref{thm:general_infinite}) is satisfied.
Similarly, $\widetilde{\bm{g}}$ is another one-hidden-layer neural network, with $D$ hidden neurons. 
{For both of $\bm g$ and $\widetilde{\bm g}$, the neural network coefficients are generated from standard normal distribution.}
For invertibility consideration, the activation function used is leaky ReLU.
The positive and negative samples are generated by \eqref{eq:pnsamples}.

\paragraph{Metric and Neural Network Settings}
We continue using MI as our evaluation metric.  The settings of $\bm h(\cdot)$ and $\phi_i(\cdot)$ are the same as those in the previous experiment.

\begin{figure}[t!]
	\centering
	\subfigure{\includegraphics[width=\linewidth]{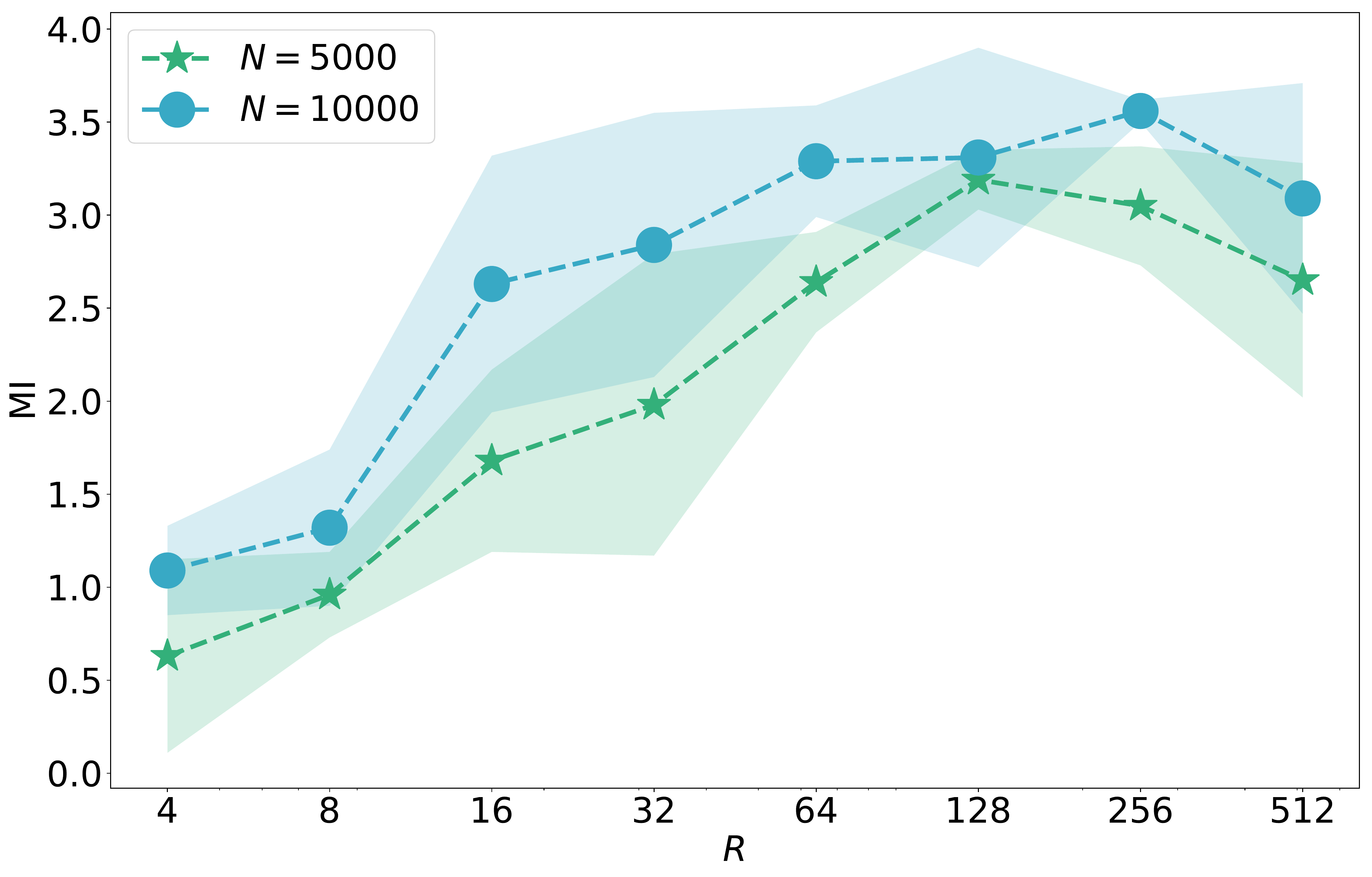}}
	\caption{The MI performance under the MVCL setting.}\label{fig:synthetic_mi_corr}
\end{figure}

\paragraph{Results}
Fig.~\ref{fig:synthetic_mi_corr} shows latent component identification performance evaluated by MI on the first view. The results are averaged over 5 random trials. 
Similar trends can be observed as in Fig.~\ref{fig:synthetic_mi_contrastive}.
That is, the performance improves when $R$ increases from 4, but starts to decline for $R\geq 128$ when $N=5000$ and $R\geq 256$ when $N=10000$.

\subsection{Real Data}
\paragraph{Data and Settings} 
In addition to synthetic data, we also use the ``EEG eye'' dataset from the UCI repository \cite{Dua:2019}. The task here is to predict whether the eyes of the subject are open or closed based on the EEG recording $\x_\ell$ at time $\ell$.
The EEG data $\x_\ell$ can be considered as a nonlinear mixture of some latent signals $\s_\ell$ ``emitted'' by the brain. 
Hence, if one could learn the unmixed latent components $\widehat{\s}_\ell=\bm{h}(\x_\ell)$ and use them as the extracted features of $\x_\ell$, it may help reduce the complexity of the learner in the downstream tasks (e.g., classification).

The data $\x_\ell$'s have fourteen dimensions. We aim to learn a five-dimensional underlying latent $\s_\ell$ for each data sample. We use 12,000 data samples as the training set to learn $\bm h(\cdot)$. Then, we train simple classifiers (i.e., SVM and logistic regression) using $\widehat{\s}_\ell=\bm h(\x_\ell)$. The classifiers are tested using 3000 test samples.
{We run the TCL framework in \cite{hyvarinen2016unsupervised} on the EEG data.}
We split the training data into 60 time frames with 200 samples within each frame. We use the frame label to be $\u_\ell$, where $\u_\ell\in\{0,\cdots,59\}$. The vector $\bm z_\ell$ is constructed following the description in Sec.~\ref{sec:aux}; also see \cite{gresele2020incomplete}.
The network structures of both $\bm{h}(\cdot)$ and $\bm{\phi}_i(\cdot)$ are as before.

\begin{figure}[t!]
	\centering
	\subfigure{\includegraphics[width=\linewidth]{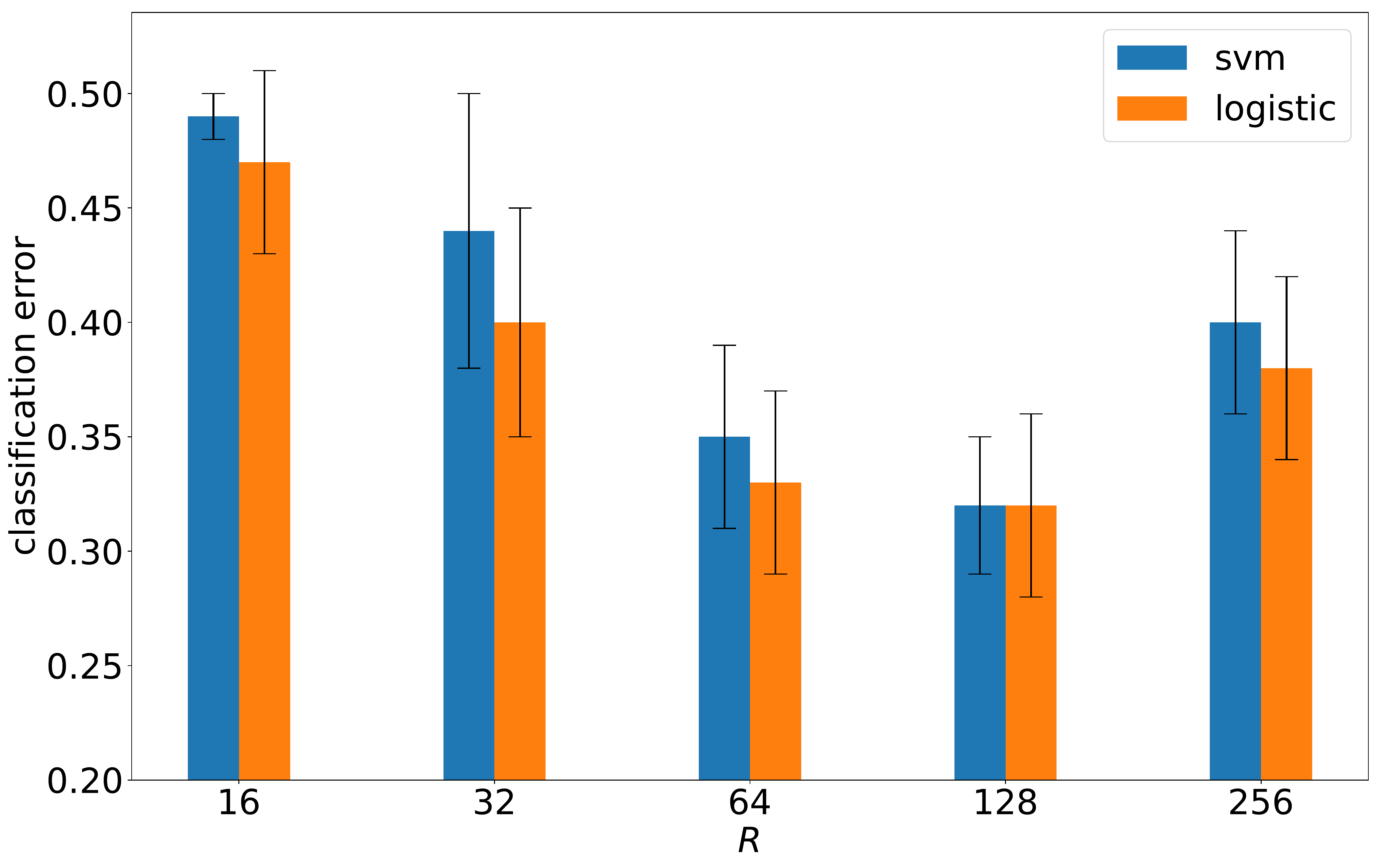}}
	\caption{Test error on the EEG data under various $R$'s.}\label{fig:eeg}
\end{figure}

\paragraph{Results} 
Fig.~\ref{fig:eeg} shows the averaged classification errors using SVM and logistic regression, respectively.
The results are averaged over 5 random trials.
One can observe a similar phenomenon as seen in the synthetic experiments. In particular, the classification performance improves as $R$ increases, because the function $\bm{h}(\cdot)$ becomes more expressive. But when the neural network gets overly complex (i.e., when $R>128$), the performance deteriorates. This result again corroborates our main result in Theorem~\ref{thm:general_finite}.

\section{Conclusion}
In this work, we investigated the identifiability problem of GCL-based nICA under a practical finite sample setting.  
The GCL-based nICA framework is an important development of the long-existing nonlinear latent component identification problem, yet existing works all used an infinite sample assumption to establish model identifiability.
Our work is the first to address the identifiability problem under a finite sample setting, to our best knowledge.
The proposed analytical framework is a nontrivial integration of properties of the logistic loss, the classic generalization theory of supervised learning, and numerical differentiation. {Unlike the existing GCL-based nICA works that all assume the use of universal exact function learners to establish identifiability}, our analysis also provides insights into the trade-off between the expressiveness and the complexity of the employed function approximators. We envision that the analytical framework can be further applied to a wider range of unsupervised/self-supervised learning problems for studying finite-sample identifiability.

{\bf Acknowledgement.} This work is supported in part by the National Science Foundation (NSF) CAREER Award ECCS-2144889, and in part by the Army Research Office (ARO) under Project ARO W911NF-21-1-0227. We also thank the anonymous reviewers for their valuable comments and constructive suggestions.

\bibliography{refs}

\begin{thebibliography}{49}
\providecommand{\natexlab}[1]{#1}
\providecommand{\url}[1]{\texttt{#1}}
\expandafter\ifx\csname urlstyle\endcsname\relax
  \providecommand{\doi}[1]{doi: #1}\else
  \providecommand{\doi}{doi: \begingroup \urlstyle{rm}\Url}\fi

\bibitem[{ Comon} \& Jutten(2010){ Comon} and Jutten]{Common2010}
{ Comon}, P. and Jutten, C.
\newblock \emph{Handbook of Blind Source Separation}.
\newblock Elsevier, 2010.

\bibitem[Arora et~al.(2012)Arora, Ge, Moitra, and
  Sachdeva]{sanjeev2012provable}
Arora, S., Ge, R., Moitra, A., and Sachdeva, S.
\newblock Provable {ICA} with unknown {G}aussian noise, with implications for
  {G}aussian mixtures and autoencoders.
\newblock In \emph{Advances in Neural Information Processing Systems},
  volume~25, 2012.

\bibitem[Arora et~al.(2019)Arora, Khandeparkar, Khodak, Plevrakis, and
  Saunshi]{arora2019theoretical}
Arora, S., Khandeparkar, H., Khodak, M., Plevrakis, O., and Saunshi, N.
\newblock A theoretical analysis of contrastive unsupervised representation
  learning.
\newblock In \emph{Proceedings of the 36th International Conference on Machine
  Learning}, 2019.

\bibitem[Bartlett \& Mendelson(2002)Bartlett and
  Mendelson]{bartlett2002rademacher}
Bartlett, P.~L. and Mendelson, S.
\newblock Rademacher and {G}aussian complexities: Risk bounds and structural
  results.
\newblock \emph{Journal of Machine Learning Research}, 3\penalty0
  (Nov):\penalty0 463--482, 2002.

\bibitem[Bengio et~al.(2013)Bengio, Courville, and
  Vincent]{bengio2013representation}
Bengio, Y., Courville, A., and Vincent, P.
\newblock Representation learning: A review and new perspectives.
\newblock \emph{IEEE Transactions on Pattern Analysis and Machine
  Intelligence}, 35\penalty0 (8):\penalty0 1798--1828, 2013.

\bibitem[Chen et~al.(2018)Chen, Li, Grosse, and Duvenaud]{chen2018isolating}
Chen, R.~T., Li, X., Grosse, R.~B., and Duvenaud, D.~K.
\newblock Isolating sources of disentanglement in variational autoencoders.
\newblock In \emph{Advances in Neural Information Processing Systems}, pp.\
  2610--2620, 2018.

\bibitem[Chen et~al.(2020)Chen, Kornblith, Norouzi, and Hinton]{chen2020simple}
Chen, T., Kornblith, S., Norouzi, M., and Hinton, G.
\newblock A simple framework for contrastive learning of visual
  representations.
\newblock In \emph{International Conference on Machine Learning}, pp.\
  1597--1607. PMLR, 2020.

\bibitem[{Comon}(1994)]{Comon1994}
{Comon}, P.
\newblock Independent component analysis, a new concept?
\newblock \emph{Signal Processing}, 36\penalty0 (3):\penalty0 287 -- 314, 1994.

\bibitem[Dua \& Graff(2017)Dua and Graff]{Dua:2019}
Dua, D. and Graff, C.
\newblock {UCI} machine learning repository, 2017.
\newblock URL \url{http://archive.ics.uci.edu/ml}.

\bibitem[Golowich et~al.(2018)Golowich, Rakhlin, and Shamir]{golowich2018size}
Golowich, N., Rakhlin, A., and Shamir, O.
\newblock Size-independent sample complexity of neural networks.
\newblock In \emph{Conference On Learning Theory}, pp.\  297--299. PMLR, 2018.

\bibitem[Goodfellow et~al.(2014)Goodfellow, Pouget-Abadie, Mirza, Xu,
  Warde-Farley, Ozair, Courville, and Bengio]{goodfellow2014generative}
Goodfellow, I., Pouget-Abadie, J., Mirza, M., Xu, B., Warde-Farley, D., Ozair,
  S., Courville, A., and Bengio, Y.
\newblock Generative adversarial nets.
\newblock In \emph{Advances in Neural Information Processing Systems}, pp.\
  2672--2680, 2014.

\bibitem[Gresele et~al.(2019)Gresele, Rubenstein, Mehrjou, Locatello, and
  Sch{\"o}lkopf]{gresele2020incomplete}
Gresele, L., Rubenstein, P.~K., Mehrjou, A., Locatello, F., and Sch{\"o}lkopf,
  B.
\newblock The incomplete {R}osetta stone problem: Identifiability results for
  multi-view nonlinear {ICA}.
\newblock In \emph{Uncertainty in Artificial Intelligence}, pp.\  217--227,
  2019.

\bibitem[Gutmann \& Hyv{\"a}rinen(2010)Gutmann and
  Hyv{\"a}rinen]{gutmann2010noise}
Gutmann, M. and Hyv{\"a}rinen, A.
\newblock Noise-contrastive estimation: A new estimation principle for
  unnormalized statistical models.
\newblock In \emph{Proceedings of the 13th International Conference on
  Artificial Intelligence and Statistics}, pp.\  297--304. JMLR Workshop and
  Conference Proceedings, 2010.

\bibitem[HaoChen et~al.(2021)HaoChen, Wei, Gaidon, and Ma]{haochen2021provable}
HaoChen, J.~Z., Wei, C., Gaidon, A., and Ma, T.
\newblock Provable guarantees for self-supervised deep learning with spectral
  contrastive loss.
\newblock In \emph{Advances in Neural Information Processing Systems}, 2021.

\bibitem[Hassoun et~al.(1995)]{hassoun1995fundamentals}
Hassoun, M.~H. et~al.
\newblock \emph{Fundamentals of artificial neural networks}.
\newblock MIT press, 1995.

\bibitem[He et~al.(2020)He, Fan, Wu, Xie, and Girshick]{he2020momentum}
He, K., Fan, H., Wu, Y., Xie, S., and Girshick, R.
\newblock Momentum contrast for unsupervised visual representation learning.
\newblock In \emph{Proceedings of the IEEE/CVF Conference on Computer Vision
  and Pattern Recognition}, pp.\  9729--9738, 2020.

\bibitem[Higgins et~al.(2017)Higgins, Matthey, Pal, Burgess, Glorot, Botvinick,
  Mohamed, and Lerchner]{higgins2016beta}
Higgins, I., Matthey, L., Pal, A., Burgess, C., Glorot, X., Botvinick, M.,
  Mohamed, S., and Lerchner, A.
\newblock {B}eta-{VAE}: Learning basic visual concepts with a constrained
  variational framework.
\newblock In \emph{International Conference on Learning Representations}, 2017.

\bibitem[Hornik(1991)]{hornik1991approximation}
Hornik, K.
\newblock Approximation capabilities of multilayer feedforward networks.
\newblock \emph{Neural Networks}, 4\penalty0 (2):\penalty0 251--257, 1991.

\bibitem[Hyvarinen(1999)]{Hyvarinen1999}
Hyvarinen, A.
\newblock Fast and robust fixed-point algorithms for independent component
  analysis.
\newblock \emph{IEEE Transactions on Neural Networks,}, 10\penalty0
  (3):\penalty0 626--634, 1999.

\bibitem[Hyvarinen \& Morioka(2016)Hyvarinen and
  Morioka]{hyvarinen2016unsupervised}
Hyvarinen, A. and Morioka, H.
\newblock Unsupervised feature extraction by time-contrastive learning and
  nonlinear {ICA}.
\newblock In \emph{Advances in Neural Information Processing Systems},
  volume~29, 2016.

\bibitem[Hyvarinen \& Morioka(2017)Hyvarinen and
  Morioka]{hyvarinen2017nonlinear}
Hyvarinen, A. and Morioka, H.
\newblock Nonlinear {ICA} of temporally dependent stationary sources.
\newblock In \emph{International Conference on Artificial Intelligence and
  Statistics}, volume~54, pp.\  460--469, 20--22 Apr 2017.

\bibitem[Hyv{\"a}rinen \& Oja(2000)Hyv{\"a}rinen and
  Oja]{hyvarinen2000independent}
Hyv{\"a}rinen, A. and Oja, E.
\newblock Independent component analysis: {A}lgorithms and applications.
\newblock \emph{Neural {N}etworks}, 13\penalty0 (4-5):\penalty0 411--430, 2000.

\bibitem[Hyv{\"a}rinen \& Pajunen(1999)Hyv{\"a}rinen and
  Pajunen]{hyvarinen1999nonlinear}
Hyv{\"a}rinen, A. and Pajunen, P.
\newblock Nonlinear {I}ndependent {C}omponent {A}nalysis: {E}xistence and
  uniqueness results.
\newblock \emph{Neural Networks}, 12\penalty0 (3):\penalty0 429--439, 1999.

\bibitem[Hyvarinen et~al.(2019)Hyvarinen, Sasaki, and
  Turner]{hyvarinen2019nonlinear}
Hyvarinen, A., Sasaki, H., and Turner, R.
\newblock Nonlinear {ICA} using auxiliary variables and generalized contrastive
  learning.
\newblock In \emph{International Conference on Artificial Intelligence and
  Statistics}, pp.\  859--868, 2019.

\bibitem[Khemakhem et~al.(2020)Khemakhem, Kingma, Monti, and
  Hyvarinen]{khemakhem2020variational}
Khemakhem, I., Kingma, D., Monti, R., and Hyvarinen, A.
\newblock Variational autoencoders and nonlinear {ICA}: A unifying framework.
\newblock In \emph{International Conference on Artificial Intelligence and
  Statistics}, pp.\  2207--2217. PMLR, 2020.

\bibitem[Kim \& Mnih(2018)Kim and Mnih]{kim2018disentangling}
Kim, H. and Mnih, A.
\newblock Disentangling by factorising.
\newblock In \emph{International Conference on Machine Learning}, volume~80,
  pp.\  2649--2658. PMLR, 10--15 Jul 2018.

\bibitem[Kingma \& Ba(2014)Kingma and Ba]{kingma2014adam}
Kingma, D.~P. and Ba, J.
\newblock Adam: A method for stochastic optimization.
\newblock \emph{arXiv preprint arXiv:1412.6980}, 2014.

\bibitem[Kozachenko \& Leonenko(1987)Kozachenko and
  Leonenko]{kozachenko1987sample}
Kozachenko, L. and Leonenko, N.~N.
\newblock Sample estimate of the entropy of a random vector.
\newblock \emph{Problemy Peredachi Informatsii}, 23\penalty0 (2):\penalty0
  9--16, 1987.

\bibitem[Kuhn(1955)]{kuhn1955hungarian}
Kuhn, H.~W.
\newblock The {H}ungarian method for the assignment problem.
\newblock \emph{Naval Research Logistics Quarterly}, 2\penalty0 (1-2):\penalty0
  83--97, 1955.

\bibitem[Locatello et~al.(2019)Locatello, Bauer, Lucic, Raetsch, Gelly,
  Sch{\"o}lkopf, and Bachem]{locatello2019challenging}
Locatello, F., Bauer, S., Lucic, M., Raetsch, G., Gelly, S., Sch{\"o}lkopf, B.,
  and Bachem, O.
\newblock Challenging common assumptions in the unsupervised learning of
  disentangled representations.
\newblock In \emph{International Conference on Machine Learning}, pp.\
  4114--4124. PMLR, 2019.

\bibitem[Locatello et~al.(2020)Locatello, Poole, R{\"a}tsch, Sch{\"o}lkopf,
  Bachem, and Tschannen]{locatello2020weakly}
Locatello, F., Poole, B., R{\"a}tsch, G., Sch{\"o}lkopf, B., Bachem, O., and
  Tschannen, M.
\newblock Weakly-supervised disentanglement without compromises.
\newblock In \emph{International Conference on Machine Learning}, pp.\
  6348--6359. PMLR, 2020.

\bibitem[Lyu \& Fu(2021)Lyu and Fu]{lyu2021simplex}
Lyu, Q. and Fu, X.
\newblock Identifiability-guaranteed simplex-structured post-nonlinear mixture
  learning via autoencoder.
\newblock \emph{IEEE Transactions on Signal Processing}, 69:\penalty0
  4921--4936, 2021.

\bibitem[Lyu \& Fu(2022)Lyu and Fu]{lyu2022finite}
Lyu, Q. and Fu, X.
\newblock Finite-sample analysis of deep cca-based unsupervised post-nonlinear
  multimodal learning.
\newblock \emph{IEEE Transactions on Neural Networks and Learning Systems},
  pp.\  1--7, 2022.

\bibitem[Lyu et~al.(2022)Lyu, Fu, Wang, and Lu]{lyu2022understanding}
Lyu, Q., Fu, X., Wang, W., and Lu, S.
\newblock Understanding latent correlation-based multiview learning and
  self-supervision: An identifiability perspective.
\newblock In \emph{International Conference on Learning Representations}, 2022.

\bibitem[Monti et~al.(2020)Monti, Zhang, and Hyv{\"a}rinen]{monti2020causal}
Monti, R.~P., Zhang, K., and Hyv{\"a}rinen, A.
\newblock Causal discovery with general non-linear relationships using
  non-linear {ICA}.
\newblock In \emph{Uncertainty in Artificial Intelligence}, pp.\  186--195.
  PMLR, 2020.

\bibitem[M{\o}rken(2013)]{morken2013numerical}
M{\o}rken, K.
\newblock Numerical algorithms and digital representation.
\newblock \emph{Lecture Notes for course MATINF1100 Modelling and
  Computations,(University of Oslo, Ch. 11, 2010)}, 2013.

\bibitem[Ni \& Gu(2016)Ni and Gu]{ni2016optimal}
Ni, R. and Gu, Q.
\newblock Optimal statistical and computational rates for one bit matrix
  completion.
\newblock In \emph{Artificial Intelligence and Statistics}, pp.\  426--434.
  PMLR, 2016.

\bibitem[Oja(1997)]{oja1997nonlinear}
Oja, E.
\newblock The nonlinear {PCA} learning rule in {I}ndependent {C}omponent
  {A}nalysis.
\newblock \emph{Neurocomputing}, 17\penalty0 (1):\penalty0 25--45, 1997.

\bibitem[Oord et~al.(2018)Oord, Li, and Vinyals]{oord2018representation}
Oord, A. v.~d., Li, Y., and Vinyals, O.
\newblock Representation learning with contrastive predictive coding.
\newblock \emph{arXiv preprint arXiv:1807.03748}, 2018.

\bibitem[Peters et~al.(2017)Peters, Janzing, and
  Sch{\"o}lkopf]{peters2017elements}
Peters, J., Janzing, D., and Sch{\"o}lkopf, B.
\newblock \emph{Elements of causal inference: foundations and learning
  algorithms}.
\newblock The MIT Press, 2017.

\bibitem[Shalev-Shwartz \& Ben-David(2014)Shalev-Shwartz and
  Ben-David]{shalev2014understanding}
Shalev-Shwartz, S. and Ben-David, S.
\newblock \emph{Understanding machine learning: From theory to algorithms}.
\newblock Cambridge university press, 2014.

\bibitem[Sprekeler et~al.(2014)Sprekeler, Zito, and
  Wiskott]{sprekeler2014extension}
Sprekeler, H., Zito, T., and Wiskott, L.
\newblock An extension of slow feature analysis for nonlinear blind source
  separation.
\newblock \emph{The Journal of Machine Learning Research}, 15\penalty0
  (1):\penalty0 921--947, 2014.

\bibitem[Taleb \& Jutten(1999)Taleb and Jutten]{taleb1999source}
Taleb, A. and Jutten, C.
\newblock Source separation in post-nonlinear mixtures.
\newblock \emph{IEEE Transactions on Signal Processing}, 47\penalty0
  (10):\penalty0 2807--2820, 1999.

\bibitem[Tian et~al.(2020)Tian, Krishnan, and Isola]{tian2020contrastive}
Tian, Y., Krishnan, D., and Isola, P.
\newblock Contrastive multiview coding.
\newblock In \emph{Computer Vision--ECCV 2020: 16th European Conference,
  Glasgow, UK, August 23--28, 2020, Proceedings, Part XI 16}, pp.\  776--794.
  Springer, 2020.

\bibitem[Tosh et~al.(2021)Tosh, Krishnamurthy, and Hsu]{tosh2021contrastive}
Tosh, C., Krishnamurthy, A., and Hsu, D.
\newblock Contrastive learning, multi-view redundancy, and linear models.
\newblock In \emph{Algorithmic Learning Theory}, pp.\  1179--1206. PMLR, 2021.

\bibitem[Tsai et~al.(2020)Tsai, Wu, Salakhutdinov, and Morency]{tsai2020self}
Tsai, Y.-H.~H., Wu, Y., Salakhutdinov, R., and Morency, L.-P.
\newblock Self-supervised learning from a multi-view perspective.
\newblock In \emph{International Conference on Learning Representations}, 2020.

\bibitem[Zhang \& Hyv{\"a}rinen(2009)Zhang and
  Hyv{\"a}rinen]{zhang2009identifiability}
Zhang, K. and Hyv{\"a}rinen, A.
\newblock On the identifiability of the post-nonlinear causal model.
\newblock In \emph{25th Conference on Uncertainty in Artificial Intelligence},
  pp.\  647--655, 2009.

\bibitem[Ziehe et~al.(2003)Ziehe, Kawanabe, Harmeling, and {
  M\"{u}ller}]{ziehe2003blind}
Ziehe, A., Kawanabe, M., Harmeling, S., and { M\"{u}ller}, K.-R.
\newblock Blind separation of post-nonlinear mixtures using linearizing
  transformations and temporal decorrelation.
\newblock \emph{Journal of Machine Learning Research}, 4:\penalty0 1319--1338,
  2003.

\bibitem[Zimmermann et~al.(2021)Zimmermann, Sharma, Schneider, Bethge, and
  Brendel]{zimmermann2021contrastive}
Zimmermann, R.~S., Sharma, Y., Schneider, S., Bethge, M., and Brendel, W.
\newblock Contrastive learning inverts the data generating process.
\newblock In \emph{Proceedings of the 38th International Conference on Machine
  Learning}, Proceedings of Machine Learning Research. PMLR, 18--24 Jul 2021.

\end{thebibliography}
\bibliographystyle{icml2022}

\newpage
\appendix
\onecolumn

\section{Proof of Lemma \ref{lemm:rademacher}}\label{ap:proof_rademacher}
We show the Rademacher complexity of the network plotted in Fig.~\ref{figs:net_struc}. Note that norm the output of $\|\bm{h}(\bm{x})\|_2$ is bounded by
\begin{align}
    \|\bm{h}(\bm{x})\|_2 \leq C_x \prod_{i=1}^L B_i
\end{align}
by simply using the Cauchy–Schwarz inequality. Next, assume that
\begin{align}
    |{h}_i(\bm{x})| \leq \frac{C_x \prod_{i=1}^L B_i}{\sqrt{D}}.
\end{align}

Then, for each of the $\phi_i$ network, the norm of its input is bound by
\begin{align}
    \sqrt{\frac{(C_x \prod_{i=1}^L B_i)^2}{D}+C_u^2}.
\end{align}

The Racemecher complexity of $\phi_i$ is bounded as
\begin{align}
    \sqrt{\frac{(C_x \prod_{i=1}^L B_i)^2}{D}+C_u^2} \prod_{i=1}^L B_i \sqrt{\frac{L}{N}}.
\end{align}

The final complexity of $r(\cdot)$ is the summation of the above, which is
\begin{align}
    \mathfrak{R}_N &\leq D\sqrt{\frac{(C_x \prod_{i=1}^L B_i)^2}{D}+C_u^2} \prod_{i=1}^L B_i \sqrt{\frac{L}{N}} \nonumber\\
    &\leq \left(C_x \prod_{i=1}^L B_i+\sqrt{D} C_u\right)\prod_{i=1}^L B_i \sqrt{\frac{DL}{N}}
\end{align}
where the second inequality is by $\sqrt{a+b}\leq \sqrt{a}+\sqrt{b}$ for $a>0,b>0$. \hfill $\blacksquare$

\section{Proof of Lemma~\ref{lem:r_gap}}\label{ap:proof_r_gap}
\subsection{Bound the Generalization Error of the Regression Function}

Define the following function:
\begin{align*}
    \ell(\bm{z}, d ;r) = \log(1+\exp[-d r(\bm{z})])
\end{align*}
Using the above notation,
consider the following loss function:
\begin{align}\label{eq:logistic_population}
    \min_{r}~\mathcal{L}(r)=\min_{r}~\mathbb{E}\left[\ell(\bm{z},d;r)\right]=\min_{r}~\mathbb{E}\left[\log(1+\exp[-d r(\bm{z})])\right]
\end{align}
where $r$ is the nonlinear function to learn. The finite-sample version is as follows:
\begin{align}\label{eq:logistic_sample}
    \min_{r}~\widehat{\mathcal{L}}(r)=\min_{r}~\frac{1}{N}\sum_{\ell=1}^N\ell(\bm{z}_\ell,d_\ell;r)=\min_{r}~\frac{1}{N}\sum_{\ell=1}^N \left[\log(1+\exp[-d_\ell r(\bm{z}_\ell)])\right]
\end{align}

{Note that we hope to bound the following
\begin{align}\label{eq:logistic_1}
    \mathcal{L}(\widehat{r}^\star) - \mathcal{L}(r^\star),
\end{align}
which can be rewritten as
\begin{align}
    \mathcal{L}(\widehat{r}^\star) -
    \min_{r\in\mathcal{R}}\mathcal{L}(r) + 
    \min_{r\in\mathcal{R}}\mathcal{L}(r) - \mathcal{L}(r^\star)\leq \mathcal{L}(\widehat{r}^\star) -
    \min_{r\in\mathcal{R}}\mathcal{L}(r) + \nu,
\end{align}
by using Assumption~\ref{as:mismatch}.
}

Invoking Theorem 26.5 of \cite{shalev2014understanding}, we have the following generalization error bound, with probability of at least $1-\delta$:
\begin{align}
    {\mathcal{L}(\widehat{r}^\star) -
    \min_{r\in\mathcal{R}}\mathcal{L}(r)}\leq 2\mathfrak{R}(\ell \circ \omega \circ r)+5c\sqrt{\frac{2\ln(8/\delta)}{N}}
\end{align}
where the notation ``$\circ$'' means function composition, $\mathfrak{R}(\ell \circ \omega \circ r)$ is the Rademacher complexity of the composed function $\ell \circ \omega \circ r$,
$r^\star$ is the optimal solution of \eqref{eq:logistic_population}, $\widehat{r}^\star$ the optimal solution of \eqref{eq:logistic_sample}, $\alpha$ is an upper bound of $|r(\bm{z})|$ and $c=\log(1+e^{\alpha})$, and
\begin{align*}
    \ell(r) = \log(1+\exp(-r)),\quad\quad
    \omega(r) = d r.
\end{align*}

By the properties of Rademacher complexity \cite{bartlett2002rademacher}, we have
\begin{align*}
    \mathfrak{R}(\ell \circ \omega \circ r) \leq \mathfrak{R}_N,
\end{align*}
where {$\mathfrak{R}_N$ is the Rademacher complexity of $r$ under $N$ i.i.d. samples,}
since both $\ell(\cdot)$ and $\omega(\cdot)$ are $1$-Lipschitz functions. Therefore, Eq.~\eqref{eq:logistic_1} becomes
\begin{align}
    \mathcal{L}(\widehat{r}^\star) - \mathcal{L}(r^\star)\leq 2\mathfrak{R}_N+\nu+5\log(1+e^{\alpha})\sqrt{\frac{2\ln(8/\delta)}{N}}.
\end{align}

\subsection{Bound the Distance Between the Learned Regression Function and the Optimal}
Next, we will bound the following error term 
\begin{align}
    \mathbb{E}[|\widehat{r}^\star(\bm{z})-{r}^\star(\bm{z})|^2],
\end{align}
 
First, using Lemma \ref{lemm:rsc}, we have
\begin{align}
    \frac{\gamma_\alpha}{2}|\widehat{r}^\star-r^\star|^2\leq\ell(\bm{z},d;\widehat{r}^\star)- \ell(\bm{z},d;r^\star)-\langle \nabla\ell(\bm{z},d;r^\star),\widehat{r}^\star-r^\star \rangle, 
\end{align}
where 
\[ \gamma_\alpha=\frac{e^\alpha}{(1+e^\alpha)^2}. \]

Taking expectation on both sides, we have
\begin{align}
    \frac{\gamma_\alpha}{2}\mathbb{E}[|\widehat{r}^\star-r^\star|^2]\leq\mathcal{L}(\widehat{r}^\star)- \mathcal{L}(r^\star)-\mathbb{E}\left[\langle \nabla\ell(\bm{z},d;r^\star),\widehat{r}^\star-r^\star \rangle\right].
\end{align}

We hope to show that 
\begin{align*}
    \mathbb{E}\left[\langle \nabla\ell(\bm{x},d;r^\star),\widehat{r}^\star-r^\star \rangle\right] \geq 0.
\end{align*}
Expand the left hand side, we have
\begin{align*}
    \mathbb{E}\left[\frac{d r^\star-d\widehat{r}^\star}{1+e^{d r^\star}}\right].
\end{align*}
Given the fact that 
\[ \mathcal{L}(\widehat{r}^\star) \geq \mathcal{L}(r^\star), \]
we have 
\begin{align*}
    \mathbb{E}\left[\log(1+e^{-d\widehat{r}^\star})\right]\geq\mathbb{E}\left[\log(1+e^{-d{r}^\star})\right]  \Longrightarrow     \mathbb{E}\left[\log \frac{1+e^{-d\widehat{r}^\star}}{1+e^{-d{r}^\star}}\right]\geq 0.
\end{align*}

By the Jensen's inequality, we have
\begin{align*}
    \mathbb{E}\left[ \frac{1+e^{-d\widehat{r}^\star}}{1+e^{-d{r}^\star}}\right]\geq 1.
\end{align*}
It can be re-written as
\begin{align*}
    \mathbb{E}\left[ \frac{1+e^{d{r}^\star}+e^{d{r}^\star-d\widehat{r}^\star}-1}{1+e^{d{r}^\star}}\right]\geq 1,
\end{align*}
which is
\begin{align*}
    \mathbb{E}\left[ \frac{e^{d{r}^\star-d\widehat{r}^\star}-e^0}{1+e^{d{r}^\star}}\right]\geq 0.
\end{align*}
Since $e^x$ is monotonic, the above implies that
\begin{align*}
    \mathbb{E}\left[\frac{d r^\star-d\widehat{r}^\star-0}{1+e^{d r^\star}}\right]\geq 0 \Rightarrow \mathbb{E}\left[\langle \nabla\ell(\bm{z},d;r^\star),\widehat{r}^\star-r^\star \rangle\right] \geq 0.
\end{align*}

Thus, we have the following inequality:
\begin{align}
    \frac{\gamma_\alpha}{2}\mathbb{E}[|\widehat{r}^\star-r^\star|^2]\leq\mathcal{L}(\widehat{r}^\star)- \mathcal{L}(r^\star)\leq 2\mathfrak{R}_N+\nu+5c\sqrt{\frac{2\ln(8/\delta)}{N}},
\end{align}
which is
\begin{align}
    \mathbb{E}[|\widehat{r}^\star(\bm{z})-r^\star(\bm{z})|^2]\leq \frac{(1+e^\alpha)^2}{e^\alpha} \left(2\mathfrak{R}_N+\nu+5c\sqrt{\frac{2\ln(8/\delta)}{N}}\right).
\end{align}

This concludes the proof. \hfill $\blacksquare$

\section{Proof of Theorem \ref{thm:general_finite}}\label{ap:proof_main}

Let us first recall the optimal $r^\star(\x,\u)$'s expression in the unlimited sample case from  \cite{hyvarinen2019nonlinear}.
Given a two-class classification problem, where the probabilities of seeing the positive class and negative class are $p(\x)$ and $q(\x)=1-p(\x)$, respectively.
We train a classifier $D$ using the following loss function:
\begin{align}\label{ap_eq:nll_loss}
    L(D) = \mathbb{E}_{\x\sim P} \left[ - \log D(\x) \right] + \mathbb{E}_{\x\sim Q} \left[ - \log (1 - D(\x)) \right]
\end{align}
where $D(\x)$ is supposed to learn the probability of the positive class that is generated following the distribution $P$ and $1-D(\x)$ the probability of class with distribution $Q$.

As shown in \cite{goodfellow2014generative}, the optimal solution is of the discriminator is as follows
\begin{align*}
    D(\x) = \frac{p(\x)}{p(\x) + q(\x)}.
\end{align*}
The above can be re-written as \cite{hyvarinen2019nonlinear}:
\begin{align*}
    D(\x) &= \frac{1}{1 + q(\x)/p(\x)}\\
    &= \frac{1}{1 + \exp ( - \log (p(\x)/q(\x))) } 
\end{align*}

In our case, we have two classes of data, i.e., $(\x_\ell,\u_\ell)\sim P=\mathbb{P}_{\x,\u}$ and  $(\x_\ell,\widetilde{\u}_\ell)\sim Q=\mathbb{P}_{\x}\mathbb{P}_{\u}$.
Note that $Q=\mathbb{P}_{\x}\mathbb{P}_{\u}$ because the latter was sampled from $\x$ and $\u$ randomly and independently.

Note that our discriminator is constructed as
\[D(\x,\u) = \frac{1}{1+\exp(-r(\x,\u))},\]
following the standard logistic regression formulation.
Hence, the optimal $r(\x,\u)$ can be written as
\begin{align}\label{ap_eq:log_pdf_diff}
    r^\star(\bm{x}, \bm{u}) &= \log \left( p(\bm{x}, \bm{u}) / p(\bm{x}) p(\bm{u})\right) \\
    &= \log p(\bm{x} | \bm{u})  - \log p(\bm{x}). \nonumber
\end{align}

\subsection{Cross-Derivative Estimation Using Numerical Derivative}
In this subsection, we estimate of the cross-derivative using the generalization bound above. First, using Lemma~\ref{lem:r_gap}, we define 
\begin{align*}
    \varepsilon = \frac{(1+e^\alpha)^2}{e^\alpha} \left(2\mathfrak{R}_N+\nu+5c\sqrt{\frac{2\ln(8/\delta)}{N}}\right).
\end{align*}

We denote
\begin{align*}
    \varepsilon_\ell = \left(\widehat{r}^\star\left(\bm{z}_\ell\right)-r^\star\left(\bm{z}_\ell\right)\right)^2
\end{align*}
as a realization of $\varepsilon$.
Using these notations {and the i.i.d. assumption}, we have
\[ \mathbb{E}_{\z_\ell\sim {\cal D}}[\varepsilon_\ell] \leq \varepsilon, \]
where ${\cal D}$ stands for the distribution that generates $
\z_1,\ldots,\z_N$.

Note the learned regression function $\widehat{r}^\star(\cdot)$ is constructed as follows:
\begin{align}\label{eq:rhat}
    \widehat{r}^\star(\bm{x},\u) = \sum_{i=1}^D \phi_i^\star\left(h_i^\star \left(\bm{x}\right),\bm{u}\right)
\end{align}
where $\bm{h}^\star=(h_1^\star,\cdots,h_D^\star)$ is invertible and smooth. 

Recall that the optimal regression function $r^\star(\bm{z})$ can be written as the difference of log PDF [cf. Eq.~\eqref{ap_eq:log_pdf_diff}]:
\begin{align}\label{eq:rstar}
    r^\star(\bm{z}) &= \log p(\bm{x} | \bm{u}) + \log p(\bm{u}) - (\log p(\bm{x}) + \log p(\bm{u})) \\
    &=\left(\sum_{i=1}^D q_i(f_i(\bm{x}),\bm{u}) + \log p(\bm{u}) + \log \det \bm{J} \right) - \left(\log p_s(\bm{f}(\bm{x})) + \log p(\bm{u}) + \log \det \bm{J}\right) \nonumber\\
    &= \sum_{i=1}^D q_i(f_i(\bm{x}),\bm{u}) - \log p_s(\bm{f}(\bm{x})) \nonumber
\end{align}
where {$p_s$ is the distribution of $\s$,}
$\bm{J}$ is the Jacobian of $\bm{f}=\bm{g}^{-1}$ and the related terms are cancelled.
Also recall that we have defined the relations:
\begin{align}\label{eq:yv}
\bm{y} =\bm{h}(\bm{x}),\quad\bm{v}(\bm{y}) =\bm{f}(\bm{h}^{-1}(\bm{y}))=\bm{s}.
\end{align}
Then, using \eqref{eq:rhat} and \eqref{eq:rstar} and the notations in \eqref{eq:yv}, we have the following
\begin{align*}
    \varepsilon_\ell = \left(\sum_{i=1}^D q_i(v_i(\bm{y}_\ell),\bm{u}_\ell) - \log p_s(\bm{v}(\bm{y}_\ell))-\sum_{i=1}^D \phi_i\left([\bm{y}_\ell]_i,\bm{u}_\ell\right)\right)^2
\end{align*}
with $\mathbb{E}_{ {\cal D}}[\varepsilon_\ell]\leq \varepsilon$.

\subsubsection{Recall the Proof of the Population Case}\label{ap:proof_infinite}
To understand our development, we first recall the key steps in the proof of the infinite sample case from \cite{hyvarinen2019nonlinear}.

{\bf Step 1.} By training the regression function defined in \eqref{eq:regression}, we have the optimal solution under unlimited infinite samples equivalent to the log PDF difference as shown in \eqref{ap_eq:log_pdf_diff} 
\begin{align}
    \underbrace{\sum_{i=1}^D \phi^\star_i\left(h^\star_i\left(\bm{x}\right),\bm{u}\right)}_{\widehat{r}^\star(\bm{x},\bm{u})} = \underbrace{\log p(\bm{x} | \bm{u})  - \log p(\bm{x})}_{{r}^\star(\bm{x},\bm{u})}.
\end{align}

{\bf Step 2.} The equation from Step 1 can be further expanded by following \eqref{eq:rstar}, i.e.,
\begin{align}\label{ap_eq:core_equation}
    \sum_{i=1}^D \phi^\star_i\left(y_i,\bm{u}\right) = \sum_{i=1}^D q_i(v_i(\bm{y}),\bm{u}) - \log p_s(\bm{v}(\bm{y})).
\end{align}

{\bf Step 3.} Denote $\eta(\bm{y})=\log p_s(\bm{v}(\bm{y}))$. Taking derivatives w.r.t. $y_j$ and $y_k$ on \eqref{ap_eq:core_equation}, we have
\begin{align}\label{ap_eq:linear_system}
    \sum_{i} \left(q''_{i}\frac{\partial {v}_i(\bm{y})}{\partial y_j} \frac{\partial {v}_i(\bm{y})}{\partial y_k} + q'_{i}\frac{\partial^2 {v}_i(\bm{y})}{\partial y_j \partial y_k}\right) - \frac{\partial^2 \eta(\bm{y})}{\partial y_j \partial y_k} = 0,
\end{align}
since the term on the LHS of Eq.~\eqref{ap_eq:core_equation} is gone.

By putting together 
\[ \bm{\kappa}_{jk}=\left[\frac{\partial {v}_1(\bm{y})}{\partial y_j} \frac{\partial {v}_i(\bm{y})}{\partial y_k}, \cdots, \frac{\partial {v}_1(\bm{y})}{\partial y_j} \frac{\partial {v}_i(\bm{y})}{\partial y_k}, \frac{\partial^2 {v}_1(\bm{y})}{\partial y_j \partial y_k}, \cdots, \frac{\partial^2 {v}_D(\bm{y})}{\partial y_j \partial y_k}\right]^\top \]
as a single vector, there could be different versions of Eq.~\eqref{ap_eq:linear_system} since the coefficients $q_i''$ and $q_i'$ are determined by $\bm{u}$.
Suppose for $\bm{u}_0$ we have
\begin{align}\label{ap_eq:u0}
    \sum_{i} \widetilde{q}''_{i}\frac{\partial {v}_i(\bm{y})}{\partial y_j} \frac{\partial {v}_i(\bm{y})}{\partial y_k} + \widetilde{q}'_{i}\frac{\partial^2 {v}_i(\bm{y})}{\partial y_j \partial y_k} - \frac{\partial^2 \eta(\bm{y})}{\partial y_j \partial y_k}=0.
\end{align}

Then, for another $\u_1,\cdots,\u_{2D}$ we have $2D$ different versions of \eqref{ap_eq:linear_system}. By subtracting \eqref{ap_eq:u0} from the $2D$ equations and using the assumption of Variability, we have
\begin{align}
    \bm{W}\bm{\kappa}_{jk}=\bm{0}
\end{align}
where $\bm{W}$ is full column rank.

Therefore, we can reach the conclusion of Theorem \ref{thm:general_infinite} by using {Fact~\ref{lemm:gamma}}.\hfill $\blacksquare$

\subsubsection{The Finite-Sample Case}
Unlike the population case, one cannot directly establish the cross-derivative equations for any point $\bm y$. Instead, we can estimate the corresponding quantity of $\phi(\bm y_\ell)$ at any observed point $\bm y_\ell$ using numerical differentiation techniques.
{
Similar to \eqref{ap_eq:core_equation}, we look at the finite sample version defined as
}
\begin{align}
    {t(\y_\ell)} = \sum_{i=1}^D q_i(v_i(\bm{y}_\ell),\bm{u}_\ell) - \log p_s(\bm{v}(\bm{y}_\ell))-\sum_{i=1}^D \phi^\star_i\left([\bm{y}_\ell]_i,\bm{u}_\ell\right),
\end{align}
of which we hope to numerically estimate its cross-derivative, i.e., $\frac{\partial^2 t(\y)}{\partial y_j\partial y_k}$. {Note that under population case, $t(\bm{y}_\ell)=0$ for any $\ell$.}

Next, we define:
\begin{align*}
    \Delta \bm{y}_{jk}^{++}&=[\bm 0,\ldots,+\Delta y_j,\ldots, \bm 0,\ldots,+\Delta y_k,\ldots, \bm 0]^\top,\\
    \Delta \bm{y}_{jk}^{+-}&=[\bm 0,\ldots,+\Delta y_j,\ldots, \bm 0,\ldots,-\Delta y_k,\ldots, \bm 0]^\top,\\
    \Delta \bm{y}_{jk}^{-+}&=[\bm 0,\ldots,-\Delta y_j,\ldots, \bm 0,\ldots,+\Delta y_k,\ldots, \bm 0]^\top,\\
    \Delta \bm{y}_{jk}^{--}&=[\bm 0,\ldots,-\Delta y_j,\ldots, \bm 0,\ldots,-\Delta y_k,\ldots, \bm 0]^\top,
\end{align*}
with $\Delta y_j>0$ and $\Delta y_k>0$ for any $j,k\in[D]$ with $j < k$.

Define $\y_{\widehat{\ell}}=\y_\ell + \Delta \y_{jk}^{++}$, $\y_{\widetilde{\ell}}=\y_\ell + \Delta \y_{jk}^{+-}$, $\y_{\overline{\ell}}=\y_\ell + \Delta \y_{jk}^{-+}$, and $\y_{{\ell}'}=\y_\ell + \Delta \y_{jk}^{--}$.
Then, we have
\begin{align*}
    \varepsilon_{\widehat{\ell}} &= \left(\sum_{i=1}^D q_i(v_i(\bm{y}_{\widehat{\ell}}),\bm{u}_\ell) - \log p_s(\bm{v}(\bm{y}_{\widehat{\ell}}))-\sum_{i=1}^D \phi^\star_i\left([\bm{y}_{\widehat{\ell}}]_i,\bm{u}_\ell\right)\right)^2,\\
    \varepsilon_{\widetilde{\ell}} &= \left(\sum_{i=1}^D q_i(v_i(\bm{y}_{\widetilde{\ell}}),\bm{u}_\ell) - \log p_s(\bm{v}(\bm{y}_{\widetilde{\ell}}))-\sum_{i=1}^D \phi^\star_i\left([\bm{y}_{\widetilde{\ell}}]_i,\bm{u}_\ell\right)\right)^2,\\
    \varepsilon_{\overline{\ell}} &= \left(\sum_{i=1}^D q_i(v_i(\bm{y}_{\overline{\ell}}),\bm{u}_\ell) - \log p_s(\bm{v}(\bm{y}_{\overline{\ell}}))-\sum_{i=1}^D \phi^\star_i\left([\bm{y}_{\overline{\ell}}]_i,\bm{u}_\ell\right)\right)^2,\\
    \varepsilon_{{\ell}'} &= \left(\sum_{i=1}^D q_i(v_i(\bm{y}_{{\ell}'}),\bm{u}_\ell) - \log p_s(\bm{v}(\bm{y}_{{\ell}'}))-\sum_{i=1}^D \phi^\star_i\left([\bm{y}_{{\ell}'}]_i,\bm{u}_\ell\right)\right)^2.
\end{align*}
where $\bm{u}_\ell$ remains the same. {Note there exist such points $\z_{\hat{\ell}}=(\x_{\hat{\ell}},\u_\ell)$, $\z_{\widetilde{\ell}}=(\x_{\widetilde{\ell}},\u_\ell)$, $\z_{\overline{\ell}}=(\x_{\overline{\ell}},\u_\ell)$ and $\z_{{\ell}'}=(\x_{{\ell}'},\u_\ell)$ in the domain of ${\cal X}
\times {\cal U}$. }

{Using numerical differentiation of multivariate function \cite{morken2013numerical}},
the cross-derivative of a function $\psi(x,y)$ can be numerically estimated as
\begin{align}
     \frac{\partial^2 \psi(x,y)}{\partial x \partial y}\approx \frac{\psi(x+\Delta x,y+\Delta y)-\psi(x+\Delta x,y-\Delta y)}{4\Delta x \Delta y}
    -\frac{\psi(x-\Delta x,y+\Delta y)-\psi(x-\Delta x,y-\Delta y)}{4\Delta x \Delta y}.
\end{align}
The exact relation between the left and right hand sides are as follows:
\begin{align*}
    \frac{\partial^2 \psi(x,y)}{\partial x \partial y}&=\frac{\psi(x+\Delta x,y+\Delta y)-\psi(x+\Delta x,y-\Delta y)}{4\Delta x \Delta y}
    -\frac{\psi(x-\Delta x,y+\Delta y)-\psi(x-\Delta x,y-\Delta y)}{4\Delta x \Delta y}\nonumber\\
    & -\frac{\Delta x^2}{6}\frac{\partial^4 \psi(\xi_{11}, \xi_{21})}{\partial x^3\partial y}-\frac{\Delta y^2}{6}\frac{\partial^4 \psi(\xi_{12}, \xi_{22})}{\partial x\partial y^3}
    -\frac{\Delta x^3}{48\Delta y}\left(\frac{\partial^4 \psi(\xi_{13},\xi_{23})}{\partial x^4}-\frac{\partial^4 \psi(\xi_{14},\xi_{24})}{\partial x^4}\right)\\
    &-\frac{\Delta x \Delta y}{8}\left(\frac{\partial^4 \psi(\xi_{15},\xi_{25})}{\partial x^2\partial y^2}-\frac{\partial^4 \psi(\xi_{16},\xi_{26})}{\partial x^2\partial y^2}\right)
    -\frac{\Delta y^3}{48\Delta x}\left(\frac{\partial^4 \psi(\xi_{17},\xi_{27})}{\partial y^4}-\frac{\partial^4 \psi(\xi_{18},\xi_{28})}{\partial y^4}\right),
\end{align*}
where $\xi_{1i}\in(x-\Delta x,x+\Delta x)$ and $\xi_{2i}\in(y-\Delta y,y+\Delta y)$ for $i\in\{1,\cdots,8\}$.

{
Denote $\eta(\bm{y}_\ell)=\log p_s(\bm{v}(\bm{y}_\ell))$. Note that the analytical form of cross derivative $\frac{\partial^2 t(\y_\ell)}{\partial y_j\partial y_k}$ is 
\begin{align}
    \frac{\partial^2 t(\y_\ell)}{\partial y_j\partial y_k}&=\sum_{i} q''_{i}\frac{\partial {v}_i(\bm{y}_\ell)}{\partial y_j} \frac{\partial {v}_i(\bm{y}_\ell)}{\partial y_k} + q'_{i}\frac{\partial^2 {v}_i(\bm{y}_\ell)}{\partial y_j \partial y_k} - \frac{\partial^2 \eta(\bm{y}_\ell)}{\partial y_j \partial y_k},
\end{align}
which can be also expressed as 
\begin{align}\label{ap_eq:cross_estimate1}
    \frac{\partial^2 t(\y_\ell)}{\partial y_j\partial y_k}&=\frac{\pm\sqrt{\varepsilon_{\widehat{\ell}}}\mp \sqrt{\varepsilon_{\widetilde{\ell}}}\mp \sqrt{\varepsilon_{\overline{\ell}}}\pm \sqrt{\varepsilon_{\ell'}}}{4\Delta y_j \Delta y_k}
    -\frac{\Delta y_j^2}{6}\frac{\partial^4 t(\bm \xi_1)}{\partial y_j^3\partial y_k}-\frac{\Delta y_k^2}{6}\frac{\partial^4 t(\bm \xi_2)}{\partial y_j\partial y_k^3}
    -\frac{\Delta y_j^3}{48\Delta y_k}\left(\frac{\partial^4 t(\bm \xi_3)}{\partial y_j^4}-\frac{\partial^4 t(\bm \xi_4)}{\partial y_j^4}\right)\\
    &-\frac{\Delta y_j \Delta y_k}{8}\left(\frac{\partial^4 t(\bm \xi_5)}{\partial y_j^2\partial y_k^2}-\frac{\partial^4 t(\bm \xi_6)}{\partial y_j^2\partial y_k^2}\right)
    -\frac{\Delta y_k^3}{48\Delta y_j}\left(\frac{\partial^4 t(\bm \xi_7)}{\partial y_k^4}-\frac{\partial^4 t(\bm \xi_8)}{\partial y_k^4}\right),\nonumber
\end{align}
where $\bm \xi_m$'s are vectors satisfying
\[ \bm \xi_m = \theta_m \y_{\widehat{\ell}}+(1-\theta_m)\y_{\ell'},\  m\in\{1,\cdots,8\}, \]
where $\theta_m\in(0,1)$.
}

Note that with a different $\bm{u}_{\widetilde{\ell}}$, we have the following
\begin{align}\label{ap_eq:cross_estimate2}
    &\quad\sum_{i} \widetilde{q}''_{i}\frac{\partial {v}_i(\bm{y}_\ell)}{\partial y_j} \frac{\partial {v}_i(\bm{y}_\ell)}{\partial y_k} + \widetilde{q}'_{i}\frac{\partial^2 {v}_i(\bm{y}_\ell)}{\partial y_j \partial y_k} - \frac{\partial^2 \eta(\bm{y}_\ell)}{\partial y_j \partial y_k}\\
    &=\frac{\pm\sqrt{\widetilde{\varepsilon}_{\widehat{\ell}}}\mp \sqrt{\widetilde{\varepsilon}_{\widetilde{\ell}}}\mp \sqrt{\widetilde{\varepsilon}_{\overline{\ell}}}\pm \sqrt{\widetilde{\varepsilon}_{\ell'}}}{4\Delta y_j \Delta y_k}
    -\frac{\Delta y_j^2}{6}\frac{\partial^4 t(\widetilde{\bm \xi}_1)}{\partial y_j^3\partial y_k}-\frac{\Delta y_k^2}{6}\frac{\partial^4 t(\widetilde{\bm \xi}_2)}{\partial y_j\partial y_k^3}
    -\frac{\Delta y_j^3}{48\Delta y_k}\left(\frac{\partial^4 t(\widetilde{\bm \xi}_3)}{\partial y_j^4}-\frac{\partial^4 t(\widetilde{\bm \xi}_4)}{\partial y_j^4}\right) \nonumber\\
    &-\frac{\Delta y_j \Delta y_k}{8}\left(\frac{\partial^4 t(\widetilde{\bm \xi}_5)}{\partial y_j^2\partial y_k^2}-\frac{\partial^4 t(\widetilde{\bm \xi}_6)}{\partial y_j^2\partial y_k^2}\right)
    -\frac{\Delta y_k^3}{48\Delta y_j}\left(\frac{\partial^4 t(\widetilde{\bm \xi}_7)}{\partial y_k^4}-\frac{\partial^4 t(\widetilde{\bm \xi}_8)}{\partial y_k^4}\right), \nonumber
\end{align}

By subtracting \eqref{ap_eq:cross_estimate2} from \eqref{ap_eq:cross_estimate1}, the $\frac{\partial^2 \eta(\bm{y}_\ell)}{\partial y_j \partial y_k}$ term is gone. Taking absolute value and {expectation w.r.t $\bm{y}$} 
\begin{align*}
    &\quad\mathbb{E}\left[\left|\sum_{i} (q''_{i}-\widetilde{q}''_{i})\frac{\partial {v}_i(\bm{y}_\ell)}{\partial y_j} \frac{\partial {v}_i(\bm{y}_\ell)}{\partial y_k} + (q'_{i}-\widetilde{q}'_{i})\frac{\partial^2 {v}_i(\bm{y}_\ell)}{\partial y_j \partial y_k}\right|\right]\\
    &\leq \frac{\mathbb{E}\left[\sqrt{{\varepsilon}_{\widehat{\ell}}}\right]+ \mathbb{E}\left[\sqrt{{\varepsilon}_{\widetilde{\ell}}}\right] + \mathbb{E}\left[\sqrt{{\varepsilon}_{\overline{\ell}}}\right] + \mathbb{E}\left[\sqrt{{\varepsilon}_{\ell'}}\right]}{4\Delta y_j \Delta y_k}
    +\frac{\mathbb{E}\left[\sqrt{\widetilde{\varepsilon}_{\widehat{\ell}}}\right]+ \mathbb{E}\left[\sqrt{\widetilde{\varepsilon}_{\widetilde{\ell}}}\right] + \mathbb{E}\left[\sqrt{\widetilde{\varepsilon}_{\overline{\ell}}}\right] + \mathbb{E}\left[\sqrt{\widetilde{\varepsilon}_{\ell'}}\right]}{4\Delta y_j \Delta y_k}\\
    &\quad + \frac{\Delta y_j^2}{3}C_t+\frac{\Delta y_k^2}{3}C_t
    +\frac{\Delta y_j^3 C_t}{12\Delta y_k}+\frac{\Delta y_j \Delta y_k C_t}{2}
    +\frac{\Delta y_k^3 C_t}{12\Delta y_j}\\
    &\leq \frac{2\sqrt{\varepsilon}}{\Delta y_j\Delta y_k} + \frac{\Delta y_j^2}{3}C_t+\frac{\Delta y_k^2}{3}C_t
    +\frac{\Delta y_j^3 C_t}{12\Delta y_k}+\frac{\Delta y_j \Delta y_k C_t}{2}
    +\frac{\Delta y_k^3 C_t}{12\Delta y_j}
\end{align*}
{where the first inequality is by triangle inequality and the assumption on the bound of fourth-order derivative, while the second inequality is by Jensen's inequality, i.e. $\sqrt{\mathbb{E}[x]}\geq \mathbb{E}[\sqrt{x}]$.}

Then, we hope to find the optimal upper bound
\begin{align*}
    \inf_{\Delta y_j,\Delta y_k}\ & \frac{2\sqrt{\varepsilon}}{\Delta y_j\Delta y_k} + \frac{\Delta y_j^2}{3}C_t+\frac{\Delta y_k^2}{3}C_t
    +\frac{\Delta y_j^3 C_t}{12\Delta y_k}+\frac{\Delta y_j \Delta y_k C_t}{2}
    +\frac{\Delta y_k^3 C_t}{12\Delta y_j}.
\end{align*}
{To find an upper bound, we let $\Delta y=\Delta y_j=\Delta y_k$, with $\Delta y>0$. Such simplification leads to a looser upper bound but easier to derive.} Then, we have the following optimization problem:
\begin{align}\label{ap_eq:general_opt_bound}
    \inf_{\Delta y}\ & \frac{2\sqrt{\varepsilon}}{\Delta y^2}+\frac{\Delta y^2}{3}C_t+\frac{\Delta y^2}{3}C_t
    +\frac{\Delta y^2 C_t}{12}+\frac{\Delta y^2 C_t}{2}
    +\frac{\Delta y^2 C_t}{12}.
\end{align}

Note that the function in \eqref{ap_eq:general_opt_bound} is convex, so the optimal is at
\begin{align*}
    \Delta y^\star = \left(\frac{36\sqrt{\varepsilon}}{C_t}\right)^{1/4}.
\end{align*}

Then, the cross-derivative can be bounded by
\begin{align*}
    \mathbb{E}\left[\left|\sum_{i} (q''_{i}-\widetilde{q}''_{i})\frac{\partial {v}_i(\bm{y}_\ell)}{\partial y_j} \frac{\partial {v}_i(\bm{y}_\ell)}{\partial y_k} + (q'_{i}-\widetilde{q}'_{i})\frac{\partial^2 {v}_i(\bm{y}_\ell)}{\partial y_j \partial y_k}\right|\right] \leq \frac{\sqrt{3C_t}\varepsilon^{1/4}}{3}.
\end{align*}

With
\[ \varepsilon = \frac{(1+e^\alpha)^2}{e^\alpha} \left(2\mathfrak{R}_N+\nu+5\log(1+e^{\alpha})\sqrt{\frac{2\ln(8/\delta)}{N}}\right), \]
we have
\begin{align}\label{ap_eq:general_bound_temp}
    \mathbb{E}\left[\left|\sum_{i} (q''_{i}-\widetilde{q}''_{i})\frac{\partial {v}_i(\bm{y}_\ell)}{\partial y_j} \frac{\partial {v}_i(\bm{y}_\ell)}{\partial y_k} + (q'_{i}-\widetilde{q}'_{i})\frac{\partial^2 {v}_i(\bm{y}_\ell)}{\partial y_j \partial y_k}\right|\right] \leq \frac{\sqrt{3C_t}(1+e^\alpha)^{1/2}}{3e^{\alpha/4}} \left(2\mathfrak{R}_N+\nu+5\log(1+e^{\alpha})\sqrt{\frac{2\ln(8/\delta)}{N}}\right)^{1/4},
\end{align}
for all pairs of $(j,k)$ with $j<k$. Thus we have $D(D-1)/2$ such inequalities above.

By the assumption of Variability \cite{hyvarinen2019nonlinear}, there exists $\bm{u}_i$ with $i\in\{0,\cdots,2D\}$, such that the matrix
\begin{align*}
    \bm{W} = [\bm{w}(\bm{y},\bm{u}_1)-\bm{w}(\bm{y},\bm{u}_0),\cdots,\bm{w}(\bm{y},\bm{u}_{2D})-\bm{w}(\bm{y},\bm{u}_0)],
\end{align*}
as defined in \eqref{eq:def_W} is full rank where $\bm{w}(\bm{y},\bm{u})$ is defined \eqref{eq:def_w_vector}. This further implies that we have $2D$ different versions of \eqref{ap_eq:general_bound_temp} with various coefficients. Putting the $2D$ inequalities together, we have the following bound
\begin{align}
    \mathbb{E}\left[\|\bm{W}\bm{\kappa}_{jk}\|_1\right]\leq
    2D \frac{\sqrt{3C_t}(1+e^\alpha)^{1/2}}{3e^{\alpha/4}} \left(2\mathfrak{R}_N+\nu+5\log(1+e^{\alpha})\sqrt{\frac{2\ln(8/\delta)}{N}}\right)^{1/4},
\end{align}
with the vector $\bm{\kappa}_{jk}$ defined earlier 
\begin{align*}
    \bm{\kappa}_{jk}=\left[
    \frac{\partial {v}_1(\bm{y}_\ell)}{\partial y_j} \frac{\partial {v}_1(\bm{y}_\ell)}{\partial y_k},\cdots,\frac{\partial {v}_D(\bm{y}_\ell)}{\partial y_j} \frac{\partial {v}_D(\bm{y}_\ell)}{\partial y_k},
    \underbrace{\frac{\partial^2 {v}_1(\bm{y}_\ell)}{\partial y_j \partial y_k},\cdots, \frac{\partial^2 {v}_D(\bm{y}_\ell)}{\partial y_j \partial y_k}}_{\bm{\gamma}^\top_{jk}}
    \right]^\top.
\end{align*}

Therefore, for any $(j,k)$ pair, we have
\begin{align*}
    \mathbb{E}\left[\|\bm{\gamma}_{jk}\|_2\right] \leq
    \mathbb{E}\left[\|\bm{\kappa}_{jk}\|_2\right] \leq
    \frac{2D\sqrt{3C_t}(1+e^\alpha)^{1/2}}{3e^{\alpha/4}{ \sigma_*^2}} \left(2\mathfrak{R}_N+\nu+5\log(1+e^{\alpha})\sqrt{\frac{2\ln(8/\delta)}{N}}\right)^{1/4},
\end{align*}
where we use the inequality $\|\bm{x}\|_2\leq \|\bm{x}\|_1$ and  $\sigma_*=\max\limits_{\W} \sigma_{\min}({\bm W})$.
This completes the proof. \hfill $\blacksquare$

\end{document}